\documentclass[journal]{IEEEtran}
\usepackage{multirow}
\usepackage{bbding}
\usepackage[pagebackref=true,breaklinks=true,colorlinks,bookmarks=false]{hyperref}

\usepackage[pdftex]{graphicx}

\ifCLASSINFOpdf
\else
\fi
\usepackage{bbding}
\newlength\savewidth\newcommand\shline{\noalign{\global\savewidth\arrayrulewidth
		\global\arrayrulewidth 1pt}\hline\noalign{\global\arrayrulewidth\savewidth}}

\begin{document}
%
\title{TJU-DHD: A Diverse High-Resolution Dataset for Object Detection}
%
%
%

\author{Yanwei Pang,
        Jiale Cao,
        Yazhao Li,
        Jin Xie,
        Hanqing Sun, 
        Jinfeng Gong
\thanks{The work is supported by the National Key R\&D Program of China (Grant No. 2018AAA0102800 and 2018AAA0102802) and National Natural Science
Foundation of China (Grant No. 61632018 and 61906131). (The corresponding author: Jiale Cao)}
\thanks{Y. Pang, J. Cao, Y. Li, J. Xie, and H. Sun are with the School of Electrical and Information Engineering and Tianjin Key Laboratory of Brain-inspired Intelligence Technology, Tianjin University, Tianjin 300072, China (E-mail: \{pyw,~connor,~lyztju,~jinxie,~hqSun\}@tju.edu.cn).}
\thanks{J. Gong is with the China Automotive Technology and Research Center Co., Ltd., Tianjin 3003000, China (E-mail:gongjinfeng@catarc.ac.cn).}
}
%
%

\markboth{IEEE Transactions on Image Processing}%
{Pang \MakeLowercase{\textit{et al.}}: TJU-DHD: A Diverse High-Resolution Dataset for Object Detection}
%



\maketitle

\begin{abstract}
Vehicles, pedestrians, and riders are the most important and interesting objects for the perception modules of self-driving vehicles and video surveillance. However, the state-of-the-art performance of detecting such important objects (esp. small objects) is far from satisfying the demand of practical systems. Large-scale, rich-diversity, and high-resolution datasets play an important role in developing better object detection methods to satisfy the demand. Existing public large-scale datasets such as MS COCO collected from websites do not focus on the
specific scenarios. Moreover, the popular datasets (\textit{e.g.,} KITTI and Citypersons) collected from the specific scenarios are limited in the number of images and instances, the resolution, and the diversity. To attempt to solve the problem, we build a diverse high-resolution dataset (called TJU-DHD). The dataset contains 115,354 high-resolution images (52\% images have a resolution of 1624$\times$1200 pixels and 48\% images have a resolution of at least 2,560$\times$1,440 pixels) and 709,330 labeled objects in total with a large variance in scale and appearance. Meanwhile, the dataset has a rich diversity in season variance, illumination variance, and weather variance. In addition, a new diverse pedestrian dataset is further built. With the four different detectors (\textit{i.e.,} the one-stage RetinaNet, anchor-free FCOS, two-stage FPN, and Cascade R-CNN), experiments about object detection and pedestrian detection are conducted. We hope that the newly built dataset can help promote the research on object detection and pedestrian detection in these two scenes.
The dataset is available at \url{https://github.com/tjubiit/TJU-DHD}.
\end{abstract}

\begin{IEEEkeywords}
Dataset, object detection, pedestrian detection, high resolution, large scale.
\end{IEEEkeywords}

%
\IEEEpeerreviewmaketitle

\section{Introduction}
Object detection aims to locate and classify objects in an image, which is a fundamental but challenging task in computer vision community. In recent few years, based on deep Convolutional Neural Networks (CNN) \cite{Krizhevsky_AlexNet_NIPS_2012,Simonyan_VGG_arxiv_2014,He_ResNet_CVPR_2016,Huang_DenseNet_CVPR_2017}, object detection has achieved great progress and started to be successfully applied to real life. Behind the technique of deep CNN, the large-scale image datasets, such as ImageNet \cite{Russakovsky_ImageNet_IJCV_2015}, PASCAL VOC \cite{Everingham_VOC_IJCV_2010}, and MS COCO \cite{Lin_COCO_ECCV_2014}, are another key to push the progress of object detection. These large-scale datasets (\textit{e.g.,} MS COCO \cite{Lin_COCO_ECCV_2014}) collected from the website do not focus on any specific scene. As a result, the detector trained on these about generic object datasets cannot achieve a very good performance in the specific application scene. It is necessary to develop a large-scale object dataset in specific scenes for specific application requirements.

\begin{figure}[!t]
\centering
\includegraphics[width=1.7in]{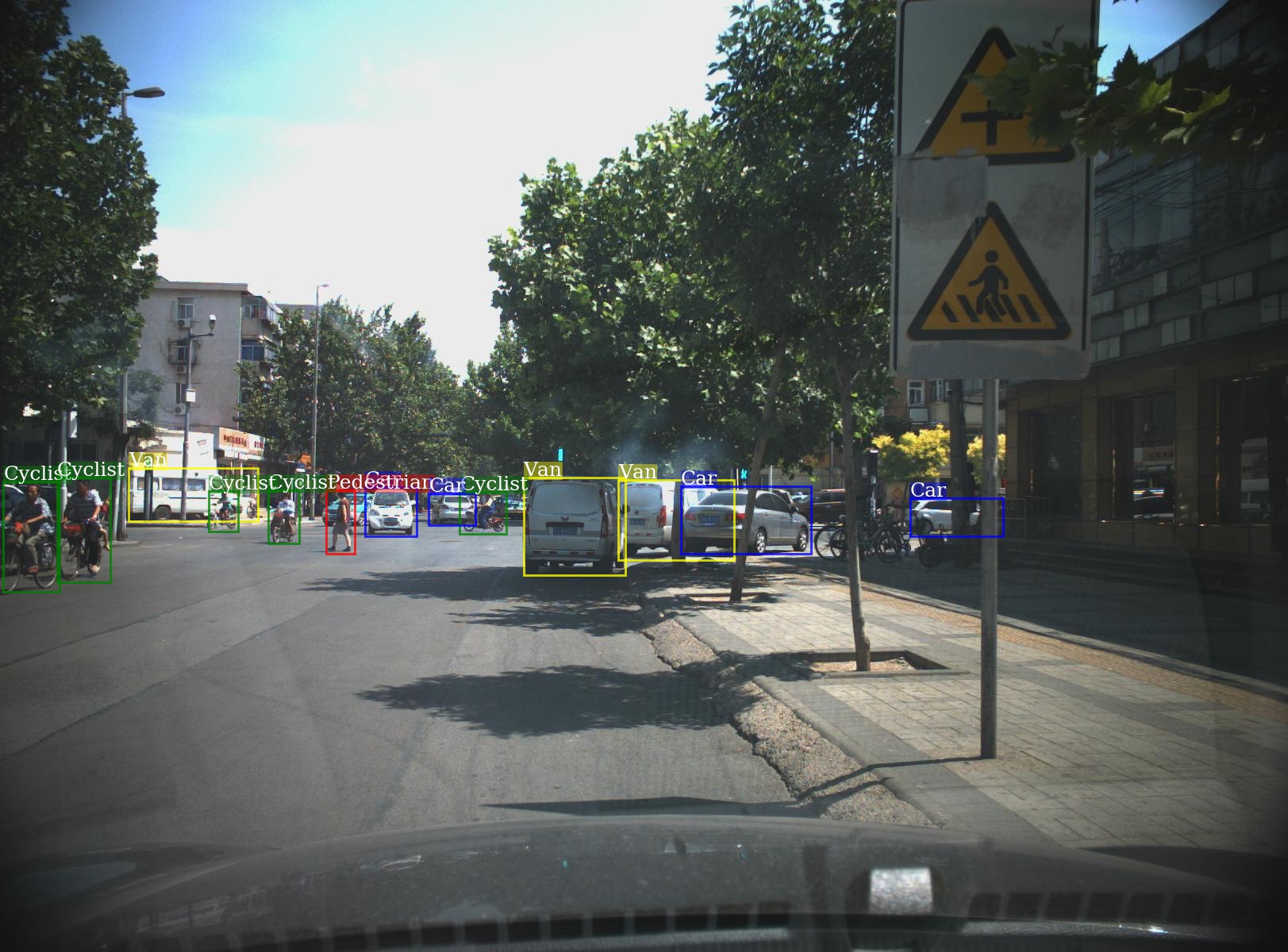}
\includegraphics[width=1.7in]{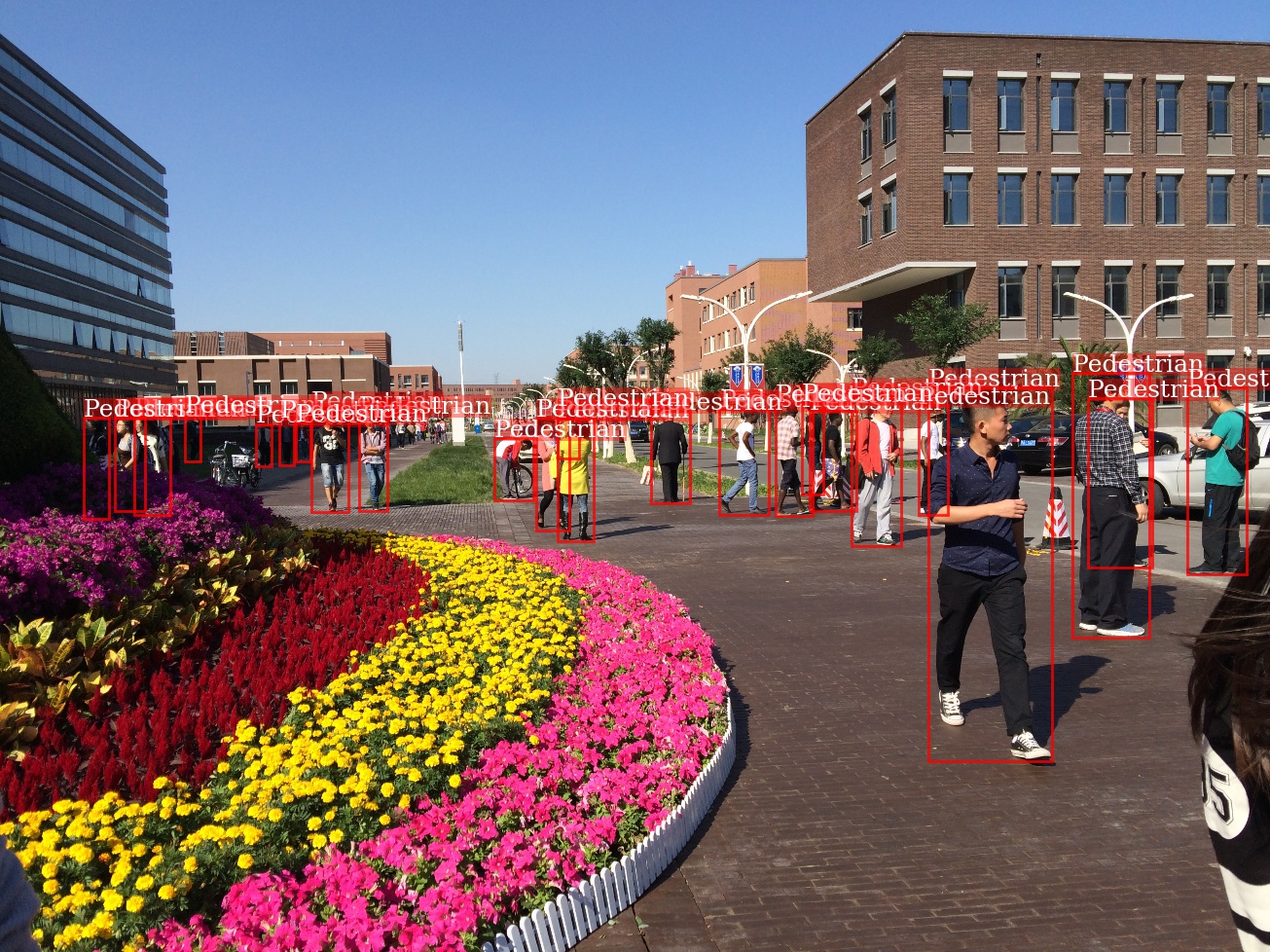}
\includegraphics[width=1.7in]{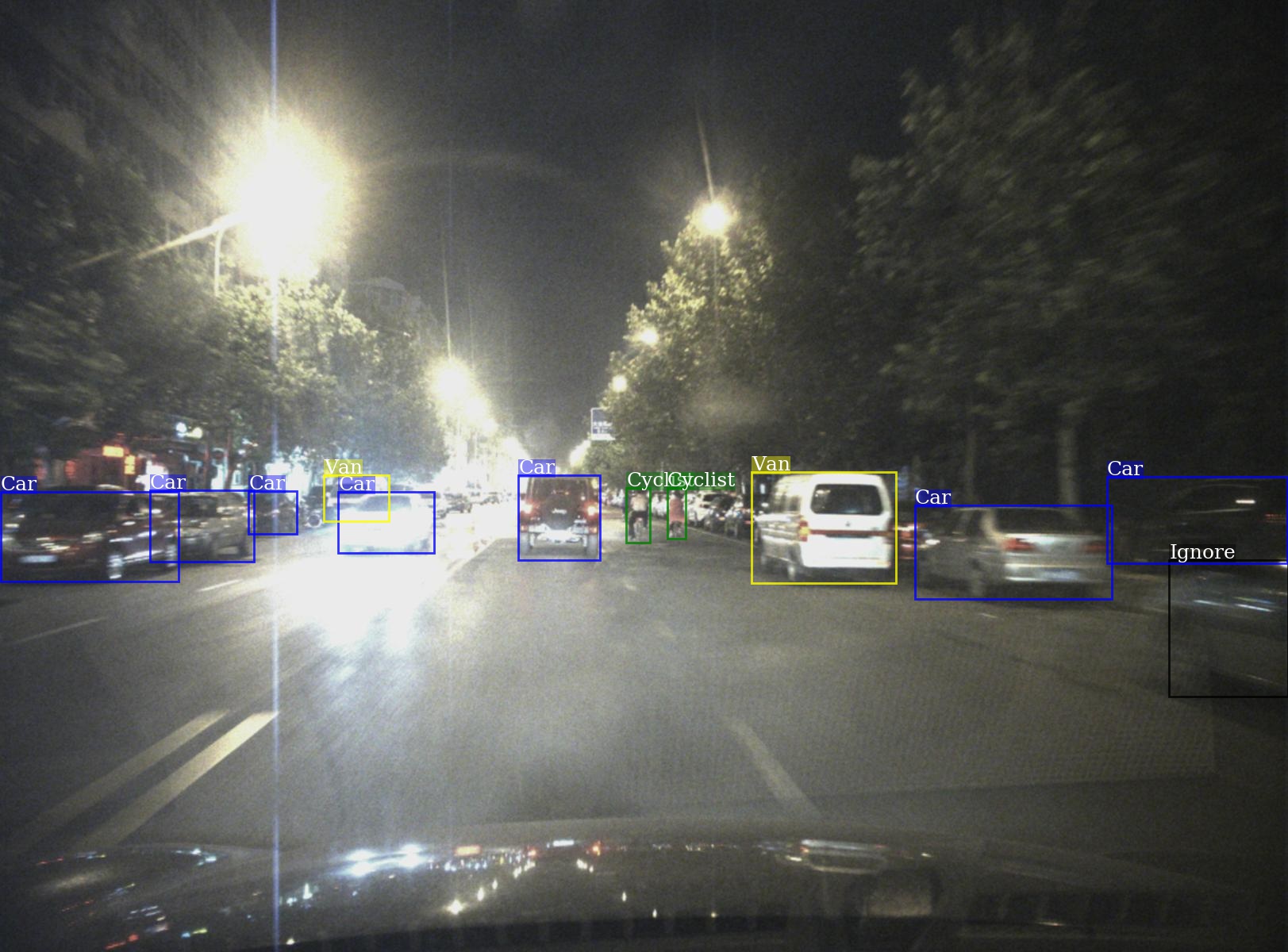}
\includegraphics[width=1.7in]{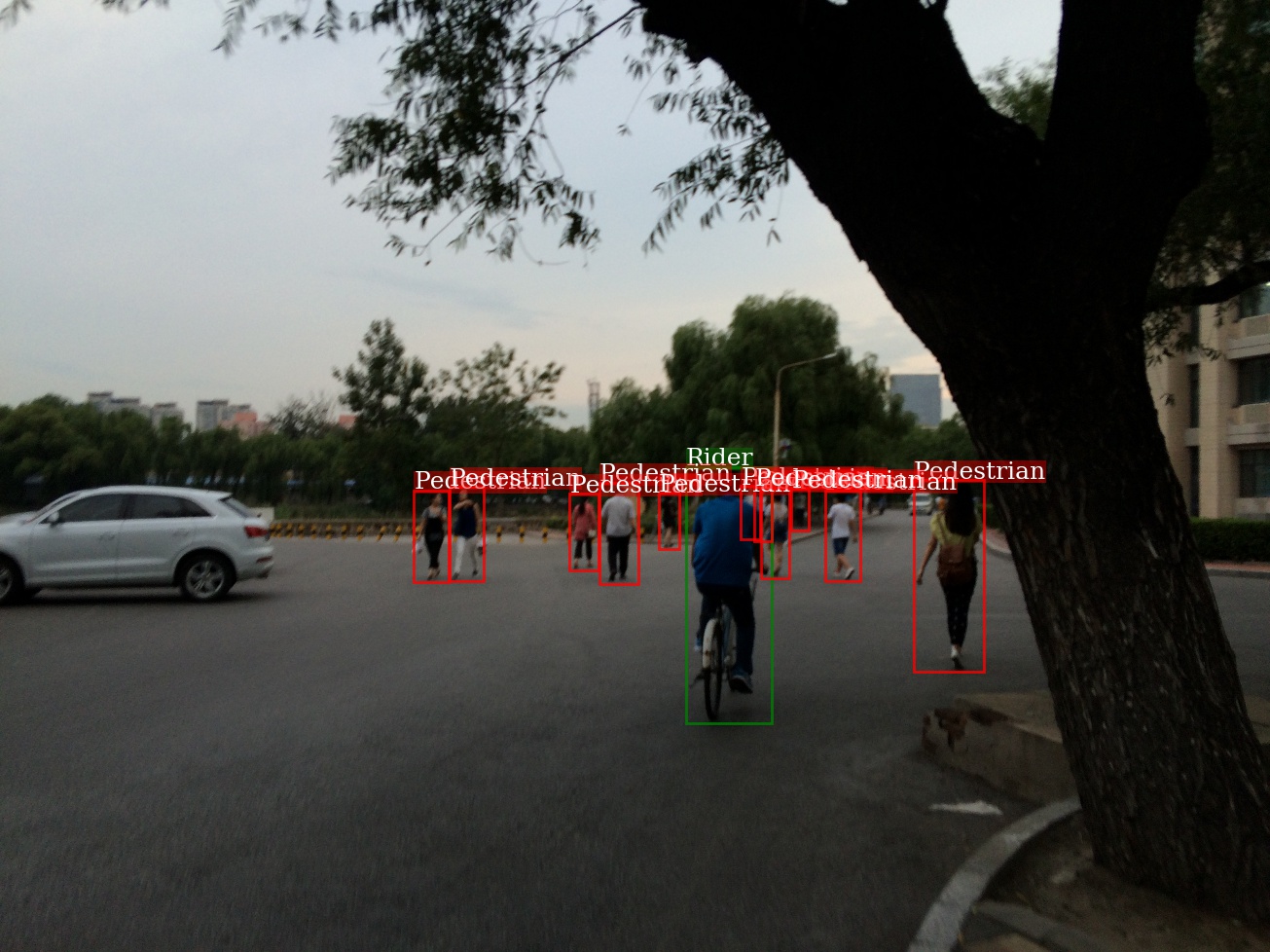}
\includegraphics[width=1.7in]{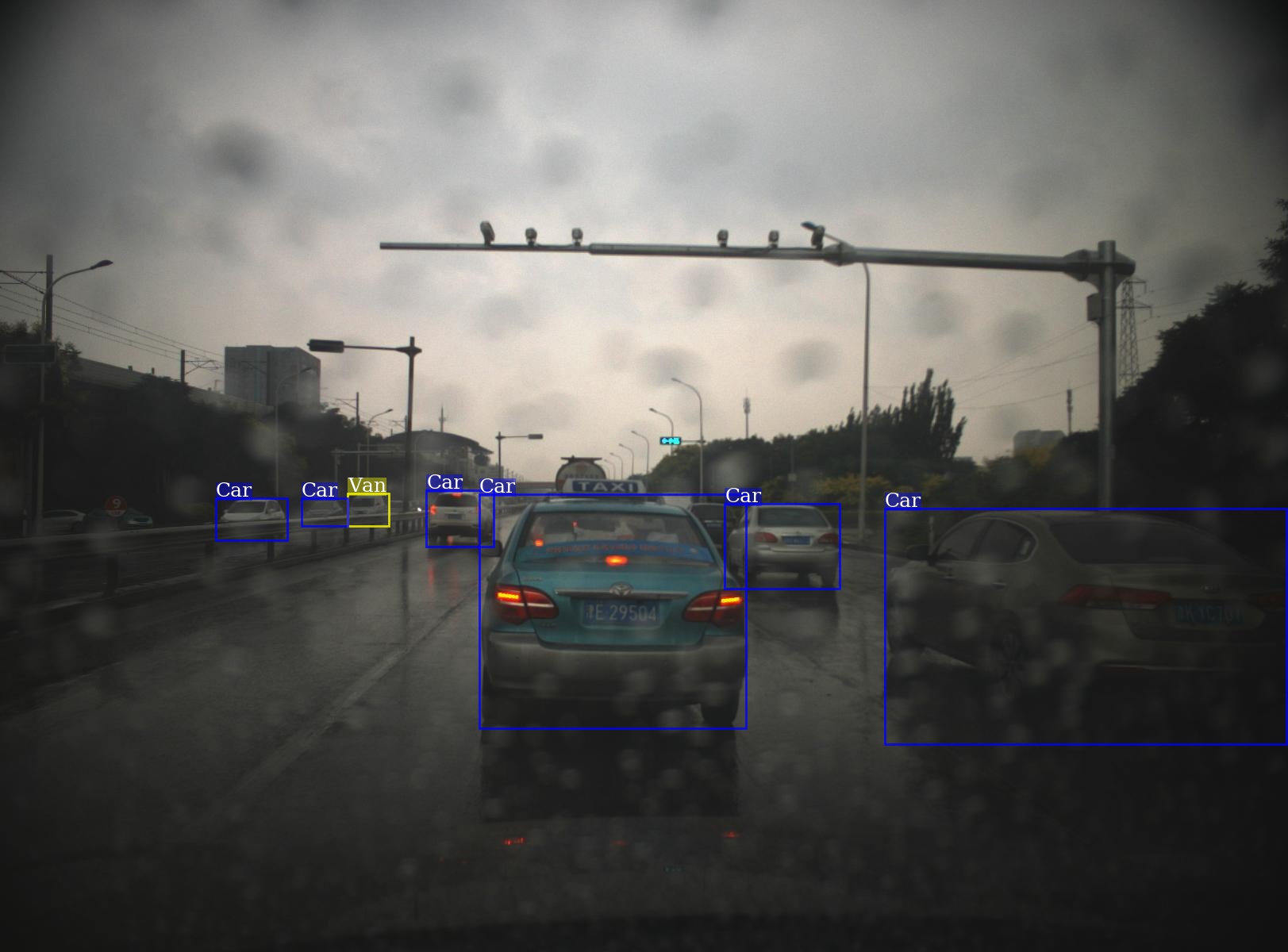}
\includegraphics[width=1.7in]{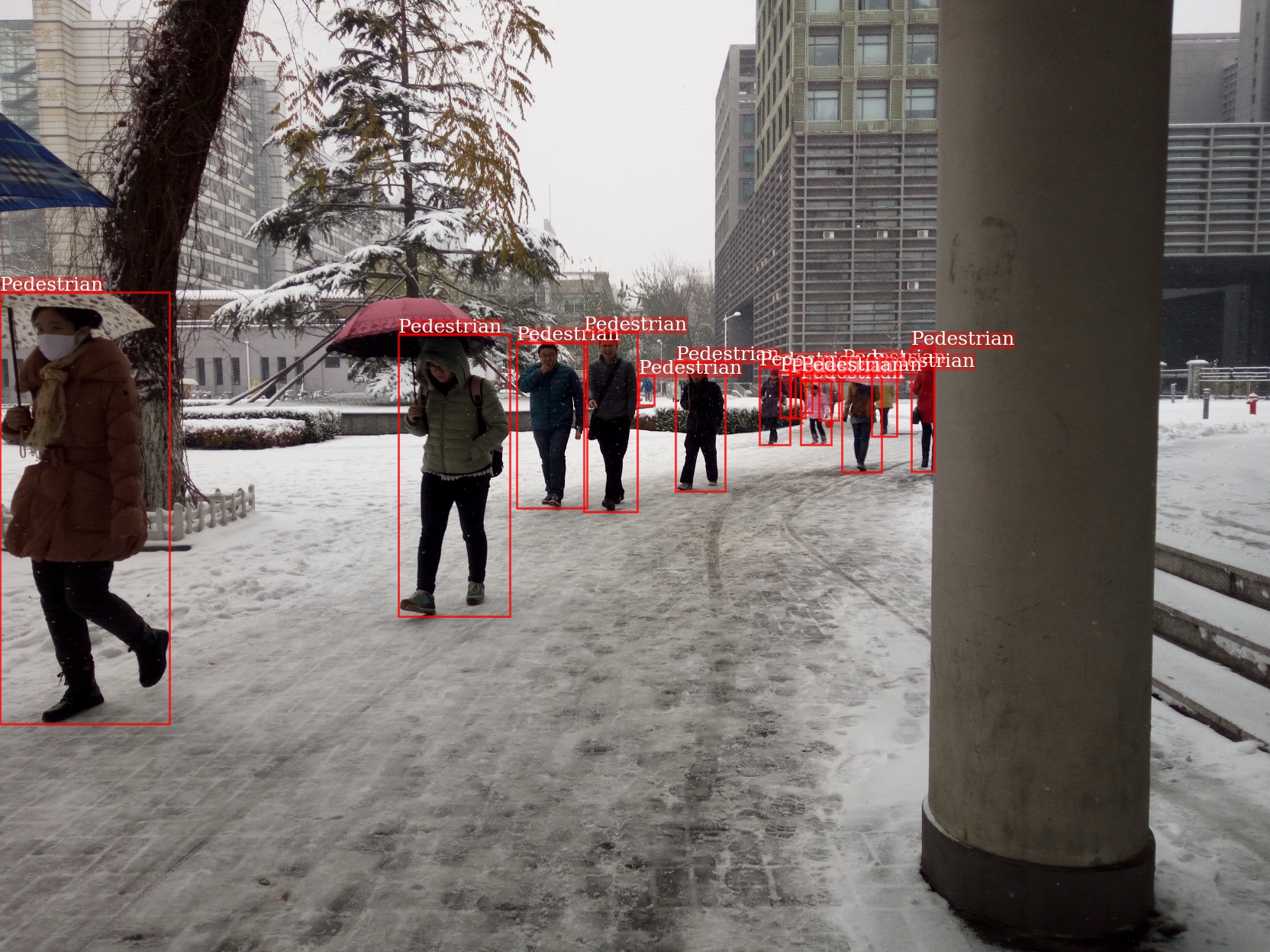}
\caption{Some examples of the newly built TJU-DHD. The dataset is collected under different scenes, different weathers, different seasons, and different illuminations. The high-resolution images in the first column are from TJU-DHD-traffic and those in the second column are from TJU-DHD-campus.}
\label{fig01}
\end{figure}

In our daily life, traffic scene and campus scene are two common scenes. It is important to detect objects (\textit{e.g.,} pedestrian and car) in these two scenes for real applications. Some datasets \cite{Dollar_PD_PAMI_2012,Geiger_KITTI_CVPR_2012,Zhang_CityPersions_CVPR_2017}  in these specific applications have been built in the past decade. For example, in the urban traffic scene, Geiger \textit{et al.} \cite{Geiger_KITTI_CVPR_2012} built the KITTI benchmark for object detection, and Doll{\'a}r \textit{et al.} \cite{Dollar_PD_PAMI_2012} proposed the Caltech dataset for pedestrian detection. The KITTI dataset \cite{Geiger_KITTI_CVPR_2012} contains 7,481 training images and 7,518 test images with a resolution of 1,240$\times$376 pixels, while the Caltech dataset \cite{Dollar_PD_PAMI_2012} has a low resolution of 640$\times$480 pixels and limited pedestrians. We argue that these two datasets are still not enough to push the progress of object detection in the specific scenes. Recently, some large-scale datasets (\textit{e.g.,} ApolloScape dataset \cite{Huang_ApooloScape_TPAMI_2019} and BDD100K \cite{Yu_BDD100K_arXiv_2019}) have been proposed. However, the ApolloScape dataset \cite{Huang_ApooloScape_TPAMI_2019} does not cover the season variance and day-night variance, and the BDD100K dataset \cite{Yu_BDD100K_arXiv_2019} has a relatively low resolution of 720p. Meanwhile, they merely focus on the traffic scene.

In this paper, we develop a new Diverse High-resolution Dataset (called TJU-DHD, collected by Tianjin university) in the traffic scene and campus scene. Fig. \ref{fig01} shows some examples. It contains 115,354 images and 709,330 labeled instances. The resolution of the images is of at least 1,624$\times$1,200 pixels and the height of the objects ranges from 11 pixels to 4,152 pixels. Meanwhile, the dataset collected over one year has a large variance in object appearance, object scale, illumination, season, and weather. As a result, the dataset has a very rich diversity. For object detection, TJU-DHD is split into two subsets: TJU-DHD-traffic subset in the traffic scene and TJU-DHD-campus subset in the campus scene. To further focus on one of the most important cases of object detection (\textit{i.e.,} pedestrian detection), we choose the images including the pedestrians in these two scenes to construct a large-scale pedestrian dataset. Based on the built dataset (\textit{i.e.,} TJU-DHD), we implement the one-stage method RetinaNet \cite{Lin_Focal_ICCV_2017}, the anchor-free method FCOS \cite{Tian_FCOS_ICCV_2019}, the two-stage method FPN \cite{Lin_FPN_CVPR_2017}, and the cascade method Cascade R-CNN \cite{Cai_Cascade_CVPR_2018} to give the baseline performance. To summarize, the contributions of this paper can be summarized as follows. 

(1) A new diverse high-resolution dataset for object detection in two important scenes is built, which has a rich variance in appearance, scale, illumination, season, and weather.

(2) A new large-scale pedestrian dataset is further proposed based on the built diverse high-resolution object dataset, which can provide both the same-scene and cross-scene evaluations. 

(3) Experiments based on four different detectors (\textit{i.e.,} one-stage RetinaNet \cite{Lin_Focal_ICCV_2017}, anchor-free FCOS \cite{Tian_FCOS_ICCV_2019}, two-stage FPN \cite{Lin_FPN_CVPR_2017}, and Cascade R-CNN \cite{Cai_Cascade_CVPR_2018}) are conducted to provide the baseline performance. We hope that this newly built high-resolution dataset can push the progress of object detection and pedestrian detection in the traffic scene and campus scene.

\section{Related work}
In this section, we firstly give a brief review of object detection datasets, including pedestrian detection datasets and face detection datasets, and secondly give a review of object detection methods, including pedestrian detection methods. 
\subsection{The datasets of object detection}
To promote the progress of object detection and provide a fair performance comparison, many datasets for object detection have been proposed in the past decade. In this subsection, we review some object detection datasets \cite{Everingham_VOC_IJCV_2010,Lin_COCO_ECCV_2014,Geiger_KITTI_CVPR_2012}, some pedestrian detection datasets \cite{Dollar_PD_PAMI_2012,Zhang_CityPersions_CVPR_2017,Shao_CrowdHuman_arxiv_2018}, and some face detection datasets \cite{Yang_WiderFace_CVPR_2016,Zhu_AFW_CVPR_2012,Jain_FDDB_TR_2010}.

\textit{Generic object detection datasets} Generally, these object datasets are usually collected from the website and do not focus on the specific application scenes. PASCAL VOC \cite{Everingham_VOC_IJCV_2010} and MS COCO \cite{Lin_COCO_ECCV_2014} are two of the most famous generic object datasets. The PASCAL VOC challenge has been held since 2006. Among these PASCAL VOC datasets, VOC2007 and VOC2012 are widely used, which have 20 object classes and over 11,000 images. Compared with PASCAL VOC, MS COCO \cite{Lin_COCO_ECCV_2014} is significantly larger in category and images, which contains 80 object classes and about 328$k$ images. Recently, some newly built datasets (\textit{e.g.,} LVIS \cite{Gupta_LVIS_CVPR_2019} and OpenImages \cite{Kuznetsova_OpenImage_arXiv_2018}) contain a very large number of object classes and images. The LVIS dataset has 1,000 object classes and 164$k$ images, and the OpenImages dataset has 600 object classes and 9.2$M$ images. Besides these generic object datasets, Some datasets, including KITTI \cite{Geiger_KITTI_CVPR_2012}, Cityscapes \cite{Cordts_Cityscapes_CVPR_2016}, Mapillary vistas \cite{Neuhold_Mapillary_ICCV_2017}, ApolloScape \cite{Huang_ApooloScape_TPAMI_2019}, and BDD100K \cite{Yu_BDD100K_arXiv_2019},  focus on the traffic scene. Specifically, the KITTI dataset has 3 classes and 14,999 images, the Cityscapes benchmark has 20 classes and 5,000 fine-annotation images, the Mapillary Vistas dataset has 66 classes and 25,000 images, and the BDD100K dataset has 10 classes and 100$k$ images.

\textit{Pedestrian detection datasets} Pedestrian detection is a key task in both self-driving and video surveillance \cite{Geiger_KITTI_CVPR_2012,Ye_DGC_TIP_2019,Xiao_JDIFL_CVPR_2017,Ye_BDCC_TIFS_2019}. The INRIA \cite{Dalal_HOG_CVPR_2005}, ETH \cite{Ess_ETH_ICCV_2007}, and Daimler \cite{Enzweiler_Daimler_TPAMI_2009} datasets are the early datasets of pedestrian detection. After those datasets, Doll{\'a}r \textit{et al.} \cite{Dollar_PD_PAMI_2012} built the larger Caltech pedestrian dataset and gave a unified evaluation of pedestrian detection. The standard Caltech pedestrian dataset consists of 4,250 images for training and 4,024 for test. However, there are only 0.32 pedestrians per image and the image resolution is low (\textit{i.e.,} $640\times480$ pixels). Recently, a new pedestrian dataset Citypersons \cite{Zhang_CityPersions_CVPR_2017} extracted from Cityscapes is proposed. There are 6.5 pedestrians per image and the image resolution is of $2,048\times1,024$ pixels. Similar to Citypersons, another new large-scale pedestrian dataset EuroCity \cite{Braun_EuroCity_PAMI_2019} is also collected in multiple European cities. To promote the progress of pedestrian detection in crowd scenes, a new dataset CrowdHuman \cite{Shao_CrowdHuman_arxiv_2018} is collected from the website, which contains about 25,000 images.

\textit{Face detection datasets} Face detection is another classic task in computer vision, which is important for face recognition and face verification. The common face detection datasets contain AFW \cite{Zhu_AFW_CVPR_2012}, FDDB \cite{Jain_FDDB_TR_2010}, PASCAL FACE \cite{Yang_PascalFace_IVC_2014}, and WiderFace \cite{Yang_WiderFace_CVPR_2016}. The AFW dataset \cite{Zhu_AFW_CVPR_2012} has 205 images with 473 faces. The FDDB dataset \cite{Jain_FDDB_TR_2010} has 2,845 images with 5,171 faces. 
The PASCAL FACE dataset \cite{Jain_FDDB_TR_2010} has 851 images with 1,341 faces. However, these datasets have the limited images and faces. To better cover face variations in pose, scale, and occlusion, the WiderFace dataset \cite{Yang_WiderFace_CVPR_2016} is proposed, which has 32,203 images with 39,3703 faces.

\subsection{The methods of object detection}
In the past decade, the dominant methods for object detection have changed from handcrafted features based ones to deep Convolutional Neural Networks (CNN) based ones. Before the CNN based methods, the researchers proposed many widely used handcrafted features (\textit{e.g.,} Haar \cite{Viola_RoFace_IJCV_2004}, LBP \cite{Ojala_LBP_PAMI_2002,Wang_HOGLBP_ICCV_2009}, ICF \cite{Dollar_ICF_BMVC_2009}, and HOG \cite{Dalal_HOG_CVPR_2005}) for object detection. With the success of deep CNN on image classification \cite{Krizhevsky_AlexNet_NIPS_2012,Simonyan_VGG_arxiv_2014,He_ResNet_CVPR_2016,Huang_DenseNet_CVPR_2017}, the researchers started to apply deep CNN to promote the progress of object detection. The CNN based methods can be mainly divided into two different classes: two-stage methods \cite{Girshick_RCNN_CVPR_2014,Girshick_FastRCNN_ICCV_2015,Ren_FasterRCNN_NIPS_2015,He_MaskRCNN_ICCV_2017,Cao_MHN_TCSVT_2019} and one-stage methods  \cite{Liu_SSD_ECCV_2016,Redmon_YOLO_CVPR_2016,Lin_Focal_ICCV_2017,Kong_RON_CVPR_2017,Nie_EFGRNet_ICCV_2019}.

Two-stage methods firstly extract some candidate object proposals and secondly classify these proposals into specific object classes. R-CNN \cite{Girshick_RCNN_CVPR_2014} is the first two-stage method, which firstly uses the handcrafted features based method selective search \cite{Uijlings_SS_IJCV_2013}  for proposal extraction, secondly calculates the CNN features for each candidate proposal, and finally classifies these proposals into specific object classes by SVM. To reduce computational costs, Fast R-CNN \cite{Girshick_FastRCNN_ICCV_2015} and SPPnet \cite{He_SPP_ECCV_2014} share the CNN feature calculations of all the candidate proposals by firstly generating the CNN features of whole image. Based on Fast R-CNN, Faster R-CNN \cite{Ren_FasterRCNN_NIPS_2015} further joins proposal extraction and proposal classification in an end-to-end network. Faster R-CNN \cite{Ren_FasterRCNN_NIPS_2015} not only improves the quality of proposal extraction but also reduces computational costs of proposal extraction. To solve the scale-variance problem in object detection, both feature pyramid methods \cite{Lin_FPN_CVPR_2017,Cai_MSCNN_ECCV_2016,Cao_MHN_TCSVT_2019,Li_TridentNet_ICCV_2019} and image pyramid methods \cite{Singh_SNIP_CVPR_2018,Singh_SNIPER_NIPS_2018,Wang_LRF_ICCV_2019,Najibi_AutoFocus_ICCV_2019} are proposed. Among the feature pyramid methods, MS-CNN \cite{Cai_MSCNN_ECCV_2016} and FPN \cite{Lin_FPN_CVPR_2017} are two representative methods, which use the shallow and high-resolution layer for small-scale object detection and use the deep and low-resolution layer for large-scale object detection. Generally, the image pyramid methods rescale the input image into a sequence of images at different scales and detect objects on each re-scaled image. To tackle the wide scale spectrum at the training stage, SNIP \cite{Singh_SNIP_CVPR_2018}  selectively back-propagates the gradients of objects at different sizes as a function of the image scale. Recently, Mask R-CNN \cite{He_MaskRCNN_ICCV_2017} is proposed to extend object detection to instance segmentation by an additional branch for mask prediction.

One-stage methods directly predict object classes and box regression offsets of dense boxes. YOLO \cite{Redmon_YOLO_CVPR_2016} and SSD \cite{Liu_SSD_ECCV_2016} are two representative one-stage methods. YOLO \cite{Redmon_YOLO_CVPR_2016} splits the input image into $N\times N$ grids and predicts object probability and class in each grid. Similar to FPN \cite{Lin_FPN_CVPR_2017}, SSD \cite{Liu_SSD_ECCV_2016} uses the shallow layer to detect small-scale objects and uses the deep layer to detect large-scale objects. Compared with two-stage methods, one-stage methods suffer more from class imbalance problem. To solve this problem, focal loss \cite{Lin_Focal_ICCV_2017} is proposed to down-weight the easy examples and up-weight the hard examples. To enhance the semantic 
of the features, many improvements \cite{Zhao_M2Det_AAAI_2018,Zhang_DES_CVPR_2018,Cao_TripleNet_CVPR_2019} have been proposed. To improve location precision, some methods \cite{Zhang_RefineDet_CVPR_2018,Jiang_IoUNet_ECCV_2018,Cao_HSD_ICCV_2019} use the cascade structure to perform the regression more than once. The above one-stage methods are anchor-based methods, which needs to set some hyper-parameters (\textit{e.g.,} anchor scales and aspect ratios).  To alleviate the drawbacks of the empirical hyper-parameters introduced by the anchor-based methods, some anchor-free methods \cite{Law_CornerNet_CVPR_2019,Zhou_CenterNet_arXiv_2019,Tian_FCOS_ICCV_2019,Yang_RepPoint_ICCV_2019,Duan_CenterNet_ICCV_2019} are recently proposed. CornerNet \cite{Law_CornerNet_CVPR_2019} uses the top-left and bottom-right corners to locate and classify objects. CenterNet \cite{Zhou_CenterNet_arXiv_2019}  predicts the center points, the heights, and the widths of objects.  FCOS \cite{Tian_FCOS_ICCV_2019}  uses a feature pyramid structure for anchor-free object detection. 

As a special and important case of object detection, pedestrian detection  \cite{Dollar_ICF_BMVC_2009,Zhou_MLL_ICCV_2017,Zhang_FasterATT_CVPR_2018,Wu_TCED_CVPR_2020,Liu_HSFD_CVPR_2019,Zhou_DFT_ICCV_2019} has also attracted much attention of researchers. Compared with generic object detection, pedestrian detection faces more severe challenges in scale variance and occlusion. Before the deep CNN based methods, the handcrafted channel features based methods \cite{Dollar_ICF_BMVC_2009,Zhang_FCF_CVPR_2015,Cao_NNNF_CVPR_2016} are dominant, which first convert the color images to ten channel images (\textit{i.e.,} three LUV channels, one gradient magnitude channel, and six oriented gradient channels) and second extract local and non-local features to learn a pedestrian detector. Recently, deep CNN based methods greatly push the progress of pedestrian detection. Some methods \cite{Mao_WHPD_CVPR_2017,Brazil_ANP_CVPR_2019,Brazil_SDS_ICCV_2017,Lin_GDFL_ECCV_2018} use semantic information to improve pedestrian detection. Some methods \cite{Song_TLL_ECCV_2018,Cao_MCF_TIP_2016,Pang_JCS_TIFS_2019,Wu_SelfMimic_ACM_2020} aim to improve small-scale pedestrian detection, while some methods \cite{Zhang_ORCNN_ECCV_2018,Zhou_Bibox_ECCV_2018,Pang_MGAN_ICCV_2019} exploit the part or visible information for occluded pedestrian detection. To improve pedestrian detection in crowded scenes, some methods \cite{Wang_Repulsion_CVPR_2018,Liu_AdaptiveNMS_CVPR_2019,Huang_R2NMS_CVPR_2020,Chi_PedHunter_AAAI_2020} exploit how to combine the highly overlapping bounding boxes.

\section{Our dataset details}
In this section, we firstly introduce the newly built diverse high-resolution dataset (called TJU-DHD) under traffic scene and campus scene, secondly describe the new large-scale pedestrian dataset under these two scenes, and finally compare the built dataset with some related datasets.

\begin{table}
\renewcommand{\arraystretch}{1.1}
\footnotesize
\centering
\caption{The number of images and instances in the training set, the validation set, and the test set of the new built dataset (called TJU-DHD). TJU-DHD contains two subsets, TJU-DHD-traffic and TJU-DHD-campus.}
\label{tab01}
\setlength{\tabcolsep}{2.5mm}{
\begin{tabular}{l|c|c|c|c}
\shline
\multirow{2}{*}{Name}       & \multicolumn{2}{|c|}{TJU-DHD-traffic} & \multicolumn{2}{|c}{TJU-DHD-campus}  \\
\cline{2-5}
& \#images &  \#instances & \#images &  \#instances \\
\shline
training set & 45,266 & 239,980 &39,727  &267,445  \\
validation set & 5,000 & 30,679 &5,204   &41,620 \\
test set & 10,000 & 60,963 &10,157   &68,643 \\
\hline
total & 60,266 & 331,622 &55,088   &377,708 \\
\shline
\end{tabular}}
\end{table}

\begin{figure}[t]
\centering
\includegraphics[width=3.5in]{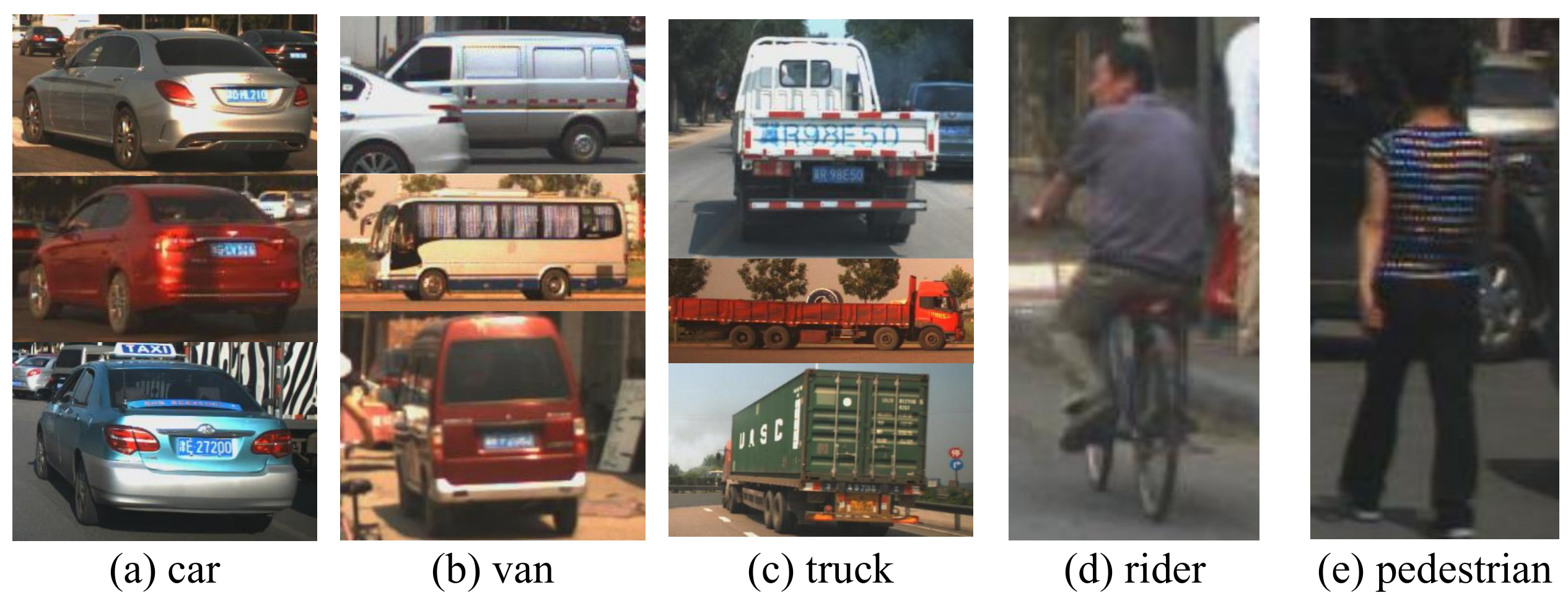}
\caption{Examples of annotated objects in five classes (\textit{i.e.,} car, van, truck, rider, and pedestrian) in the built TJU-DHD. In our dataset, sedan and SUV are treated as the same class (called car), van, minibus, and bus are all treated as the same class (called van), and  cyclist, motorcyclist, and tricyclist are all treated as the same class (called rider).}
\label{fig02}
\end{figure}

\subsection{Images, object annotations, and dataset splits}
Traffic scene and campus scene are two common scenes in our daily life. Due to the complexity of the scenes, detecting objects in these two scenes are relatively difficult. To promote the progress of object detection in these two scenes, we build a new diverse high-resolution dataset, including two different subsets corresponding to these two scenes (called TJU-DHD-traffic and TJU-DHD-campus).

TJU-DHD-traffic subset is collected by a driving car in the traffic scene, which has 60,266 images and 331,622 labeled instances (see Table \ref{tab01}). The images have a fixed high resolution  (\textit{i.e.,} 1,624$\times$1,200 pixels). We select five common and important object classes, namely car, van, truck, pedestrian, and rider, for bounding box annotations. Fig. \ref{fig02} gives some examples and illustrations of five different object classes. To annotate each object instance, a bounding box ($x_1,y_1,x_2,y_2$) represented by the left-top point ($x_1,y_1$) and the right-bottom point ($x_2,y_2$) is used. Meanwhile, the occlusion level and the truncation level of each instance are also given. The TJU-DHD-traffic is split into three sets: the training set, the validation set, and the test set. Table \ref{tab01} shows the number of images and instances in each set. Specifically, the training set contains 45,266 images and 239,980 instances, the validation set contains 5,000 images and 30,679 instances, and the test set contains 10,000 images and 60,963 instances.

TJU-DHD-campus subset is taken from multiple different mobile phones mainly on the university campus, which has 55,088 images and 377,708 labeled instances. Because the images are taken by different mobile phones, the image resolutions are high but various, which are of at least 2,560$\times$1,440 pixels. We only select two most important and common objects (\textit{i.e.,} pedestrian and rider) in campus scene for bounding box annotations. To better exploit pedestrian detection for the researchers, especially occluded pedestrian detection, we provide two bounding box annotations for each pedestrian. The two bounding boxes respectively represent the full body and the visible part of the pedestrian. The TJU-DHD-campus is also split into three different sets: the training set, the validation set, and the test set. Table \ref{tab01}  shows the number of images and instances in each set. The training set contains 39,727 images and 267,445 instances, the validation set contains 5,204 images and 41,620 instances, and the test set contains 10,157 images and 68,643 instances.

\begin{figure}[!t]
	\centering
	\includegraphics[width=1.75in]{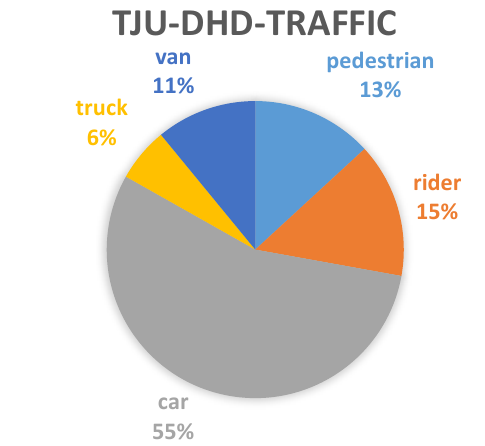}
	\includegraphics[width=1.68in]{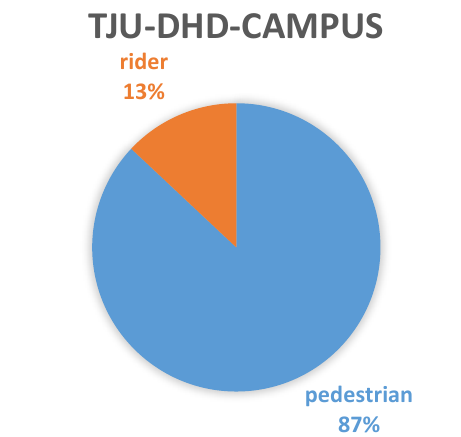}
	\caption{The proportions of different object classes.}
	\label{fig03}
\end{figure}

\begin{figure}[!t]
	\centering
	\includegraphics[width=1.7in]{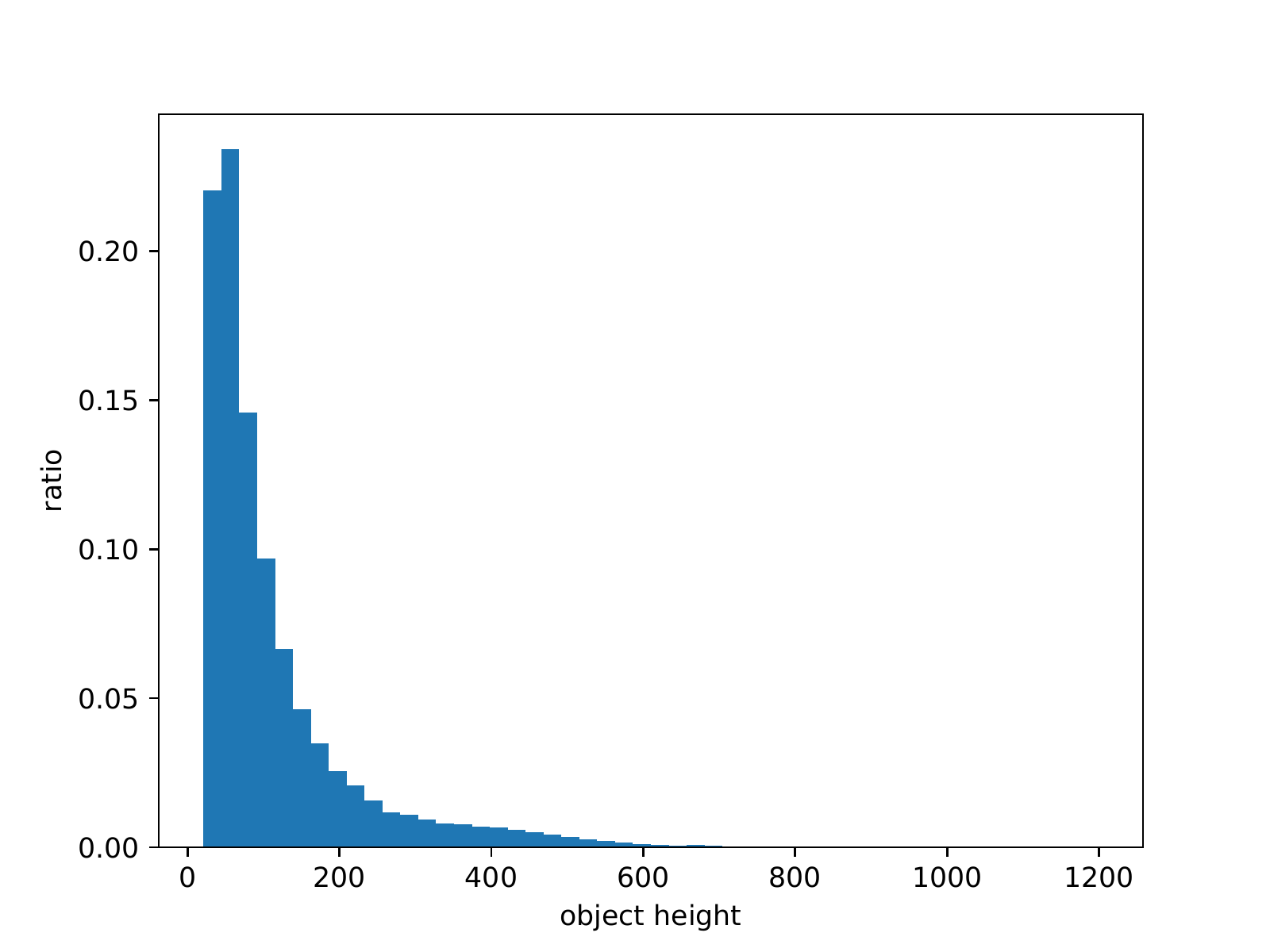}
	\includegraphics[width=1.7in]{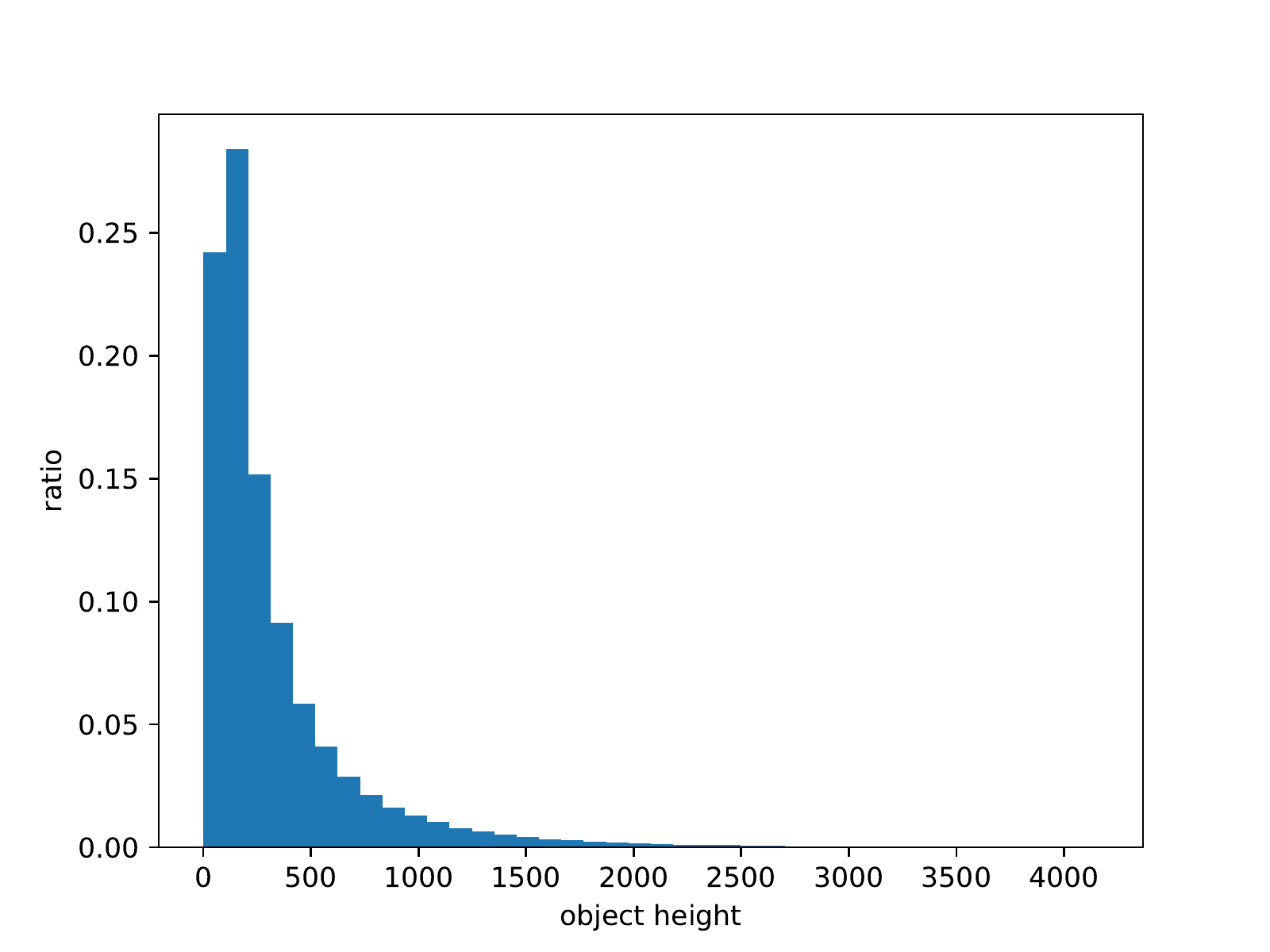}
	\caption{Scale variance in the TJU-DHD-traffic (left) and TJU-DHD-campus (right).}
	\label{fig04}
\end{figure}

\begin{figure}[!t]
	\centering
	\includegraphics[width=1.75in]{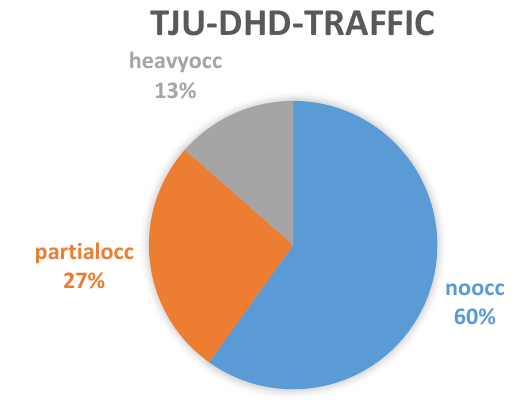}
	\includegraphics[width=1.62in]{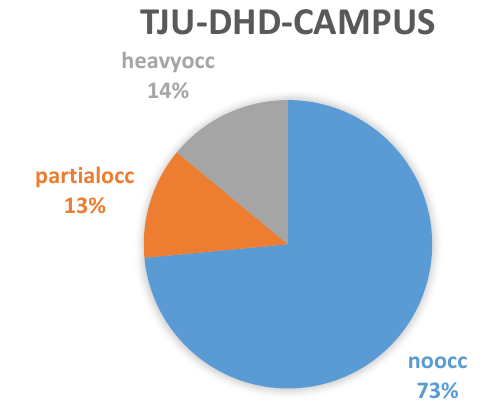}
	\caption{The proportions of different occlusion levels.}
	\label{fig05}
\end{figure}

\begin{figure*}[!t]
	\centering
	\includegraphics[width=7.2in]{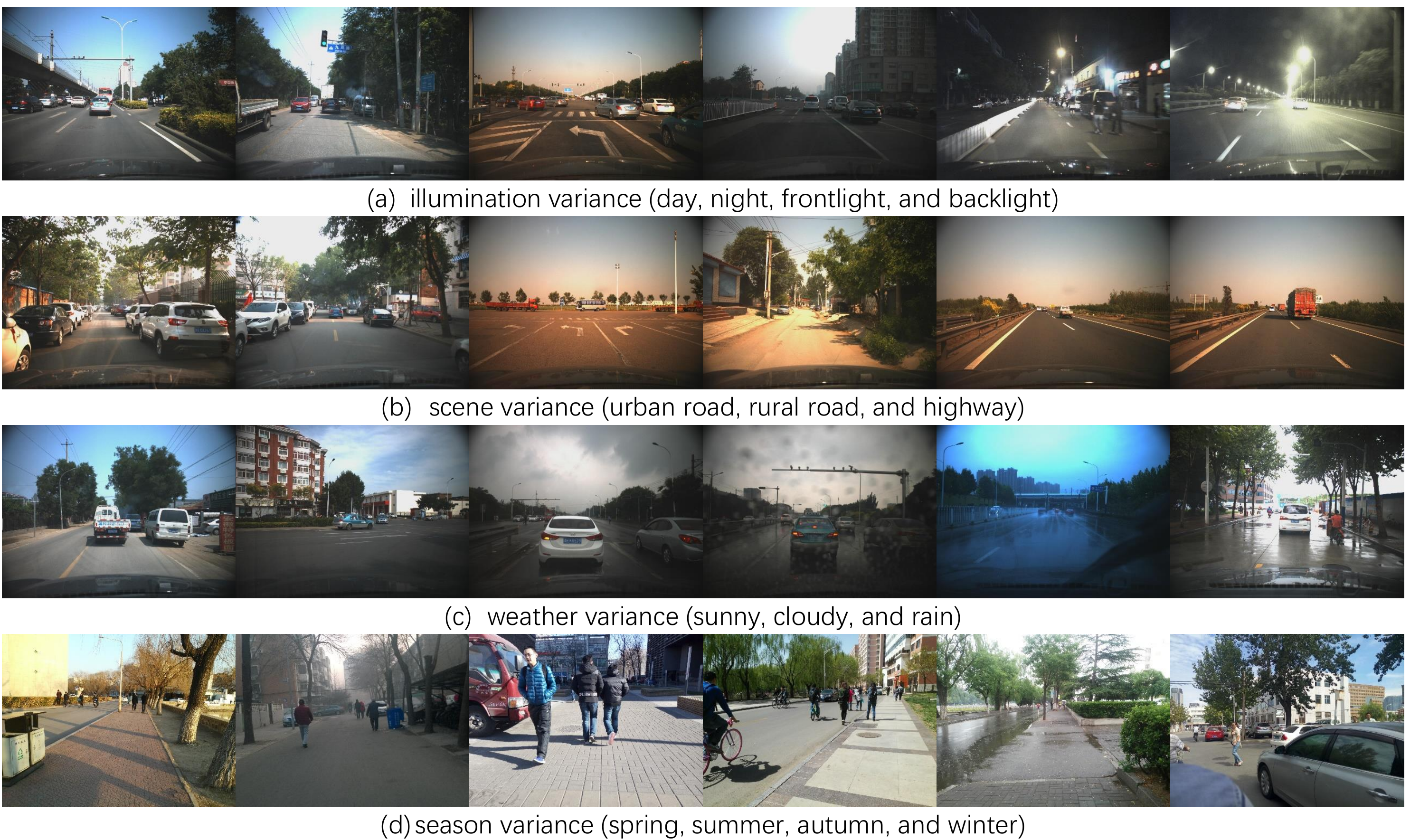}
	\caption{The rich diversity of the built TJU-DHD, which contains the variances of illumination, scene, weather, and season.}
	\label{fig06}
\end{figure*}

\subsection{Dataset statistics}
Based on the newly built dataset (called TJU-DHD), we give the detailed statistics about category variance, scale variance, occlusion variance, and dataset diversity to better know and understand this newly built dataset.

\textit{Category statistics} Fig. \ref{fig03} shows the proportions of different object classes in both the TJU-DHD-traffic and TJU-DHD-campus. In the TJU-DHD-traffic, the proportions of pedestrian, rider, car, truck, and van  are 13\%, 15\%, 55\%, 6\%, and 11\%, respectively. In the TJU-DHD-campus, the proportions of pedestrian and rider are 87\% and 13\%, respectively.

\textit{Scale statistics} Fig. \ref{fig04} shows the scale variance of objects in both the TJU-DHD-traffic and TJU-DHD-campus. we use the object height to represent its scale because the height and the scale are closely related under these two scenes. Fig. \ref{fig04}(a) shows the object height variance in the TJU-DHD-traffic. The heights vary from 20 pixels to 1,200 pixels and most objects have a height of fewer than 200 pixels. Fig. \ref{fig04}(b) shows the height variance in the TJU-DHD-campus, where the height varies from 11 pixels to 4,152 pixels. Namely, our dataset has a very large variance in object scale.

\begin{table}[t]
\renewcommand{\arraystretch}{1.1}
\footnotesize
\centering
\caption{The number of images and instances in the training set, the validation set, and the test set for pedestrian detection under two scenes.}
\label{tab03}
\setlength{\tabcolsep}{2.5mm}{
	\begin{tabular}{l|c|c|c|c}
		\shline
		\multirow{2}{*}{Name}       & \multicolumn{2}{|c|}{TJU-Ped-traffic} & \multicolumn{2}{|c}{TJU-Ped-campus}  \\
		\cline{2-5}
		& \#images &  \#instances & \#images &  \#instances \\
		\shline
		training set & 13,858 & 27,650 &39,727  &234,455  \\
		validation set & 2,136 & 5,244 &5,204   &36,161 \\
		test set & 4,344 & 10,724 &10,157   &59,007 \\
		\hline
		total & 20,338 & 43,618 &55,088   &329,623 \\
		\shline
	\end{tabular}}
\end{table}

\begin{table*}[!t]
\renewcommand{\arraystretch}{1.3}
\footnotesize
\centering
\caption{Comparisons with some related datasets. For a fair comparison, the statistics are only based on the training set. `\#avg', `vbox', and `\#ignore' represent averaged object number per image, visible box annotation, and the number of ignored bounding boxes.}
\label{tab02}
\setlength{\tabcolsep}{1.45mm}{
	\begin{tabular}{l|c|c|c|cc|cccc|c}
		\shline
		Name        & scene & day/night& seasons & \#images & resolution & \#objects & \#avg & vbox & categories    & \#ignore \\
		\shline
		Caltech \cite{Dollar_PD_PAMI_2012} & traffic& day & 1&42,782 & $640\times480$ & 13,674&  0.32 & \Checkmark & ped.  &51,092\\
		KITTI \cite{Geiger_KITTI_CVPR_2012} & traffic& day & 1 &7,481 & $1,240\times376$ &  40,570&  5.42 & \XSolidBrush &  ped., car, cyclist & 11,295\\			
		Citypersons \cite{Zhang_CityPersions_CVPR_2017} & traffic& day & 1 &2,975 & $2,048\times1,024$ & 19,238 & 6.47& \Checkmark & ped. & 6,768\\
		CrowdedHuman \cite{Shao_CrowdHuman_arxiv_2018} & crowd& day & 1 &15,000 & $>400\times300$ & 339,565&  22.64& \Checkmark &  people& 99,227\\
		EuroCity \cite{Braun_EuroCity_PAMI_2019} & traffic& day,night & 4 & 28,114& $1,920\times1,024$ & 142,736 & 5.08& \XSolidBrush &  ped., rider& 83,218\\
		\hline
		TJU-DHD-traffic & traffic& day,night & 1 & 45,266 & $1,624\times1,200$ & 239,980&  5.30& \XSolidBrush &  ped., rider, car, truck, van & 11,711\\			
		TJU-DHD-campus & campus& day,night & 4 &39,727  &$>2,560\times1,440$ &267,445&  6.73 & \Checkmark &  ped., rider &24,321\\		
		TJU-DHD-pedestrian & traffic, campus& day,night & 4 & 53,585 & $\geq 1,624\times1,200$ & 262,105&  4.89& \Checkmark  &  ped. & 73,846\\			
		\shline
	\end{tabular}}
\end{table*}

\textit{Occlusion statistics} Similar to the Caltech and Citypersons datasets \cite{Dollar_PD_PAMI_2012,Zhang_CityPersions_CVPR_2017}, object occlusion level is divided into three different levels (\textit{i.e.,} no occlusion, partial occlusion, and heavy occlusion). Partial occlusion means that the occlusion ratio $o$ is less than 0.35 and greater than 0.0 (\textit{i.e.,} $o\leq0.35$), and heavy occlusion means that the occlusion ratio $o$ is greater than 0.35 (\textit{i.e.,} $o>0.35$).  Fig. \ref{fig05} plots the proportions of different object occlusion levels in both the TJU-DHD-traffic and TJU-DHD-campus. In the TJU-DHD-traffic, the proportion of no occlusion is 60\%, the proportion of partial occlusion is 27\%, and the proportion of heavy occlusion is 13\%.  In the TJU-DHD-campus, the proportion of no occlusion is 73\%, the proportion of partial occlusion is 13\%, and the proportion of heavy occlusion is 14\%.

\textit{Diversity} The datasets are collected over more than one year. Except for the diversity in object appearance, object scale, and object density, it also covers the diversity in illumination variance, scene variance, weather variance, and season variance. Fig. \ref{fig06}(a) gives some examples of illuminate variance. Because the built dataset is captured from day to night, the illumination variance can be caused by daytime, nighttime, frontlight, and backlight. Fig. \ref{fig06}(b) gives some examples of traffic scene variance, which contains urban road, highway road, and rural road. Fig. \ref{fig06}(c) gives some examples of weather variance, which contains sunny days, cloudy days, and rainy days. Fig. \ref{fig06}(d) gives some examples of season variance from spring to winter. The season variance can cause a large variance in the appearances of both objects and background.

\subsection{The newly built pedestrian dataset}
Pedestrian detection is a typical and important case in object detection, which has drawn much attention of the researchers in the past decade. To better focus on the specific pedestrian detection, a new large-scale pedestrian dataset is built based on the TJU-DHD-traffic and TJU-DHD-campus. To simplify the expression, the newly built large-scale pedestrian dataset is called TJU-DHD-pedestrian.

We choose the images which contain the pedestrians to construct TJU-DHD-pedestrian. In TJU-DHD-pedestrian, the annotations of car, van, and truck are removed. Because the rider has a relation to the pedestrian, the annotations of rider are set as the ignored regions. As a result, TJU-DHD-pedestrian has 75,426 images, 373,241 labeled pedestrians, and 112,842 ignored bounding boxes (see Table \ref{tab03}). Instead of combining the images in two scenes together, we keep the original split of the training, validation, and test sets. Thus, TJU-DHD-pedestrian has two training, validation, and test sets. Specifically, there are 20,338 images and 43,618 labeled pedestrians in the traffic scene (called TJU-Ped-traffic), and there are 55,088 images and 329,623 labeled pedestrians in the campus scene (called TJU-Ped-campus). These two sets have a large gap in image resolution, detection scene, and instance scale. Based on these two sets, we can give a better performance analysis of the pedestrian detector with both the same-scene evaluation and the cross-scene evaluation.

\subsection{Comparison of some related datasets}
In this subsection, some related object datasets (\textit{e.g.,} Caltech \cite{Dollar_PD_PAMI_2012}, Citypersons \cite{Zhang_CityPersions_CVPR_2017}, CrowdedHuman \cite{Shao_CrowdHuman_arxiv_2018}, EuroCity \cite{Braun_EuroCity_PAMI_2019}, and KITTI \cite{Geiger_KITTI_CVPR_2012}) are compared with the newly built dataset (TJU-DHD) in Table \ref{tab02}.  For a fair comparison, the statistics are only performed on the training set. 

\textit{Caltech}\footnote{\url{http://www.vision.caltech.edu/Image_Datasets/CaltechPedestrians}} is a very popular pedestrian dataset in the past decade \cite{Dollar_PD_PAMI_2012}. The standard training set has 4,250 images. To improve detection performance, some researchers \cite{Zhang_FCF_CVPR_2015,Cao_NNNF_CVPR_2016} started to enlarge the training set by capturing every third frame. As a result, it has a large number of images (\textit{i.e.,} 42,782). However, two adjacent images are highly correlated and there are only 13,674 pedestrians. Meanwhile, the image resolution of 640$\times$480 pixels is relatively low. 

\textit{KITTI}\footnote{\url{http://www.cvlibs.net/datasets/kitti}} is a challenging benchmark \cite{Geiger_KITTI_CVPR_2012}  for the application of self-driving, which contains many different computer vision tasks (\textit{i.e.,} stereo, optical flow, visual odometry, object detection, and tracking). For object detection task, the training set has 7,481 images with a  resolution of 1,240$\times$376 pixels. Meanwhile, there are 19,238 labeled objects. Thus, the numbers of images and the instances are both limited.

\textit{Citypersons}\footnote{\url{https://bitbucket.org/shanshanzhang/citypersons}} is a recently built pedestrian dataset \cite{Zhang_CityPersions_CVPR_2017} collected from the Cityscapes benchmark \cite{Cordts_Cityscapes_CVPR_2016}. Compared with the Caltech and KITTI datasets \cite{Dollar_PD_PAMI_2012,Geiger_KITTI_CVPR_2012}, Citypersons has more pedestrians per image and larger image resolution (\textit{i.e.,} 2,048$\times$1,024 pixels). However, the numbers of images and instances are still limited.

\textit{CrowdedHuman}\footnote{\url{https://www.crowdhuman.org}} is collected from the website and aims to improve detection performance in the human crowd scene \cite{Shao_CrowdHuman_arxiv_2018}. It does not focus on the specific application scene (\textit{e.g.,} the traffic scene or the campus scene). Meanwhile, the image resolution in CrowdedHuman is relatively low.

\begin{table*}[t]
\renewcommand{\arraystretch}{1.1}
\centering
\footnotesize
\caption{Average precisions (AP) of some state-of-the-art detectors on the TJU-DHD-traffic.}
\label{tab04}	
\setlength{\tabcolsep}{3.6mm}{
\begin{tabular}{l|c|c|c|cc|ccc}
	\shline
	method   &  backbone   & input size   & AP        & AP@0.5  & AP@0.75  & AP$_s$ & AP$_m$ & AP$_l$     \\
	\shline
	(a) RetinaNet \cite{Lin_Focal_ICCV_2017}    &  ResNet50    & 1,333$\times$800    &   53.5    &  80.9  &  60.0  &  24.0 &  50.5  &  68.0  \\
	(b) FCOS \cite{Tian_FCOS_ICCV_2019}    & ResNet50    & 1,333$\times$800     &   53.8  &  80.0 &   60.1  &  24.6 &  50.6   &  68.8       \\
	(c) FPN \cite{Lin_FPN_CVPR_2017}    &  ResNet50   & 1,333$\times$800          &    55.4    & 83.4 & 63.0   &  30.4  & 52.2  &   68.2        \\			
	(d) Cascade R-CNN \cite{Cai_Cascade_CVPR_2018}    & ResNet50    & 1333$\times$800     &       57.9  & 82.7  &  66.6    & 32.6 &   54.4  &  71.4   \\
	\shline
\end{tabular}}
\end{table*}

\textit{EuroCity persons}\footnote{\url{https://eurocity-dataset.tudelft.nl}} is collected from multiple cities in Europe \cite{Braun_EuroCity_PAMI_2019}. Compared with the previous datasets \cite{Dollar_PD_PAMI_2012,Zhang_CityPersions_CVPR_2017}, EuroCity persons has a larger number of images (\textit{i.e.,} 40,219) under different illuminations from day to night. Compared with Eurocity, our proposed TJU-DHD has the following differences: (1) Our dataset is more complete, which focuses on not only pedestrian/rider detection but also vehicle detection. (2) Our pedestrian dataset is almost two times larger than EuroCity in both images and objects (see Table \ref{tab02}). Moreover, it provides visible box annotation and two different scenes (traffic and campus). (3) Our pedestrian dataset is far from saturated. In Table \ref{tab09}, the miss rate on the R set of EuroCity is 6.81\%, while miss rates on the R set of TJU-DHD-traffic and TJU-DHD-campus are 27.92\% and 22.30\%. Thus, our pedestrian dataset has more space for future research.

Overall, the advantages of our built dataset can be summarized as follows: (1) \textit{A large amount of data}. TJU-DHD-traffic, TJU-DHD-campus, TJU-DHD-pedestrian almost have the largest number of both images and instances. For example, TJU-DHD-pedestrian has almost two times larger than EuroCity in both images and pedestrians. (2) \textit{High resolution}. Our built dataset has higher resolutions than other dataset, which can provide more space for the research of small-scale object detection. TJU-DHD-traffic has a fixed high resolution of 1,624$\times$1,200 pixels, while TJU-DHD-campus has a high resolution of at least 2,560$\times$1,400 pixels. (3) \textit{Large time span}. On the one hand, TJU-DHD-traffic and TJU-DHD-campus are collected from day to night. On the other hand, TJU-DHD-campus is collected over one year. Thus, it has a rich diversity. (4) \textit{Cross-scene evaluation}. TJU-DHD-pedestrian contains the pedestrians under two different scenes, which can not only give the same-scene evaluation but also give the cross-scene evaluation.

\section{Experiments}
Recently, deep convolutional neural networks \cite{Girshick_RCNN_CVPR_2014,Girshick_FastRCNN_ICCV_2015,Ren_FasterRCNN_NIPS_2015} have achieved great success in object detection. In this section, we select four representative methods (\textit{i.e.,} RetinaNet \cite{Lin_Focal_ICCV_2017}, FCOS \cite{Tian_FCOS_ICCV_2019}, FPN \cite{Lin_FPN_CVPR_2017}, and Cascade R-CNN \cite{Cai_Cascade_CVPR_2018}) to conduct some fundamental experiments on the newly built dataset and give the baseline for future research.

\subsection{Evaluation metric}
Following the standard evaluation on the MS COCO benchmark \cite{Lin_COCO_ECCV_2014} and Caltech pedestrian dataset \cite{Dollar_PD_PAMI_2012}, mean average precision (mAP) and log-average miss rate (MR) are respectively used to evaluate object detection and pedestrian detection, which are introduced as follows.

\textit{Mean average precision} The average precision (AP) on the MS COCO benchmark is averaged under ten different intersection over union (IoU) thresholds of 0.50:0.05:0.95. Compared with AP on the PASCAL VOC \cite{Everingham_VOC_IJCV_2010}, mAP on the MS COCO considers not only the accuracy of object classification but also the precision of object location.

\textit{Log-average miss rate} It is computed by averaging log miss rates at nine different FPPI rates varying from $10^{-2}$ to $10^0$. FPPI means false positive per image. The IoU threshold used to assign a box as positive or negative is set to be 0.5.

To evaluate the detection performance under different scales, the objects are divided into different sets according to the object scales. In the TJU-DHD-traffic, 
following the same protocol as used in the MS COCO benchmark, we divide the objects  into three different sets  (\textit{i.e.,} small objects: area $<32^2$, middle objects: $32^2<$ area $<96^2$, large objects: area $>96^2$). In the TJU-DHD-campus, due to the much larger scale variations compared with the COCO benchmark, we divide the objects into four different sets according to the heights of the objects, namely tiny objects (height $<80$), small objects ($80<$ height $<160$), medium objects($160<$ height $<320$), and large objects (height $>320$). In TJU-DHD-campus dataset, there are 14.0\% tiny objects, 26.8\% small objects, 26.6\% medium objects, and 32.6\% large objects.

\subsection{Baseline detectors}
Because the feature pyramid structure is very useful for multi-scale object detection and pedestrian detection, we select four representative feature pyramid detectors to conduct the experiments on the newly built dataset (called TJU-DHD). They are the one-stage detector RetinaNet \cite{Lin_Focal_ICCV_2017}, the anchor-free detector FCOS \cite{Tian_FCOS_ICCV_2019}, the two-stage detector FPN \cite{Lin_FPN_CVPR_2017}, and the cascade detector Cascade R-CNN \cite{Cai_Cascade_CVPR_2018}. 

\textit{RetinaNet} To solve the class imbalance problem in single-stage methods (\textit{e.g.,} YOLO \cite{Redmon_YOLO_CVPR_2016} and SSD \cite{Liu_SSD_ECCV_2016}) and improve detection accuracy, RetinaNet \cite{Lin_Focal_ICCV_2017} adopts focal loss to decrease the weights of easy samples and increase the weights of hard samples. As a result, it achieves a comparable performance with two-stage methods.

\textit{FCOS} RetinaNet \cite{Lin_Focal_ICCV_2017} is an anchor-based method, which is required to design the scales and aspect ratios of anchors. Compared to RetinaNet, FCOS \cite{Tian_FCOS_ICCV_2019} is an anchor-free method, which predicts the offsets of left, right, top, and down for each point inside an object bounding box. Thus, FCOS does not require the handcrafted design of anchors.

\textit{FPN} Compared to Faster R-CNN \cite{Ren_FasterRCNN_NIPS_2015}, FPN \cite{Lin_FPN_CVPR_2017} uses the in-network layers of different resolutions to detect the objects at different scales and adopts the top-down structure to enhance the feature semantic of each output layer. Thus, FPN can largely improve the performance of object detection, especially small-scale object detection. 

\textit{Cascade R-CNN} To improve the localization accuracy, Cascade R-CNN \cite{Cai_Cascade_CVPR_2018} stacks multiple ROI detectors. The different ROI detectors are trained stage by stage with increasing IoU thresholds. As a result, Cascade R-CNN can progressively improve object localization accuracy. 

\begin{table}[t]
\renewcommand{\arraystretch}{1.1}
\centering
\footnotesize
\caption{Average precisions (AP)  per category on the TJU-DHD-traffic.}
\label{tab05}	
\setlength{\tabcolsep}{1.1mm}{
\begin{tabular}{l|c|ccccc}
	\shline
	method    & AP   & AP$_{ped}$ & AP$_{rid}$ & AP$_{car}$  & AP$_{tru}$  & AP$_{van}$     \\
	\shline
	(a) RetinaNet \cite{Lin_Focal_ICCV_2017}  &  53.5 &  58.4 &  35.9  &  48.0 &  69.7 & 55.7 \\
	(b) FCOS  \cite{Tian_FCOS_ICCV_2019}         & 53.8  &  58.7 &  45.0  &  48.4 &  71.2   &  56.0\\
	(c) FPN \cite{Lin_FPN_CVPR_2017}         &  55.4 & 60.0 &  39.7   & 50.6 &  70.9   &  55.7  \\
	(d) Cascade R-CNN \cite{Cai_Cascade_CVPR_2018}            & 57.9  & 61.7  &  42.5    & 53.7 &   73.0  &  58.7\\
	\shline
\end{tabular}}
\end{table}

\begin{table*}[t]
\renewcommand{\arraystretch}{1.1}
\centering
\footnotesize
\caption{Average precisions (AP) of some state-of-the-art detectors on the TJU-DHD-campus.}
\label{tab06}	
\setlength{\tabcolsep}{3.6mm}{
	\begin{tabular}{l|c|c|c|cc|cccc}
		\shline
		method   &  backbone   & input size   & AP        & AP@0.5  & AP@0.75  & AP$_t$   & AP$_s$   & AP$_m$   & AP$_l$     \\
		\shline
		(a) RetinaNet \cite{Lin_Focal_ICCV_2017}   &  ResNet50    & 1,333$\times$800    &48.4 &73.9 &52.4 &4.7 &27.3 &56.2 &73.8\\
		(b) FCOS \cite{Tian_FCOS_ICCV_2019}    &  ResNet50   & 1,333$\times$800         &49.3 &73.8 &53.8 &5.6 &29.6 &55.9 &74.3\\
		(c) FPN \cite{Lin_FPN_CVPR_2017}    & ResNet50    & 1,333$\times$800          &52.4 &77.5 &58.4 &8.5 &37.4 &58.6 &74.9\\
		(d) Cascade R-CNN \cite{Cai_Cascade_CVPR_2018}    & ResNet50    & 1,333$\times$800 &55.1 &77.6 &60.9 &10.8&40.1 &61.2 &78.0\\
		\shline
	\end{tabular}}
\end{table*}

\subsection{Implementation details}
These four detectors are implemented based on the open source object detection toolbox mmdetection\footnote{\url{https://github.com/open-mmlab/mmdetection}}. The widely used deep residual model ResNet50 \cite{He_ResNet_CVPR_2016} is used as the backbone. In the object detection experiments, the typical input with the resolution of 1,333$\times$800 pixels on the COCO benchmark is used for training and testing. In the pedestrian detection experiments, the typical input with the resolution of 2,048$\times$1,024 pixels on the Citypersons is used for training and testing. We use 2 NVIDIA GPUs for training. A mini-batch has 4 images per GPUs. The initial learning rate is 0.01 for FPN, 0.005 for RetinaNet, 0.005 for FCOS, and 0.01 for Cascade R-CNN. After that, the learning rate decreases at epoch 8 and epoch 11 by a factor of 10. During the inference stage, the top 100 detection bounding boxes per image are saved for performance evaluation.

\subsection{Experimental results on the TJU-DHD-traffic}
In this subsection, the experiments on TJU-DHD-traffic are conducted. Table \ref{tab04} shows the average precisions (AP) of these four detectors (\textit{i.e.,} RetinaNet \cite{Lin_Focal_ICCV_2017}, FCOS \cite{Tian_FCOS_ICCV_2019}, FPN \cite{Lin_FPN_CVPR_2017}, and Cascade R-CNN \cite{Cai_Cascade_CVPR_2018}). It can be concluded as follows: (1) The anchor-free FCOS is a little superior (+0.3\%) to the anchor-based RetinaNet and inferior (-1.6\%) to the two-stage FPN. (2) The two-stage FPN is 1.9\% better than one-stage RetinaNet. On small-scale object detection, FPN is 6.4\% better than RetinaNet.  (3) Cascade structure is very useful for accurate object detection. For example, Cascade R-CNN is even 0.7\% worse than FPN on AP@0.5, while it is 3.6\% better than FPN on AP@0.75.

To see the performance on each object category, Table \ref{tab05} further shows the AP per object category. Cascade R-CNN steadily outperforms the other three methods on each category, which further shows the effectiveness of cascade structure. The four detectors all have a relatively lower performance on the categories of car and rider. The reason may be explained as follows: (1) riders have a large inter-class variance because it contains cyclist, motorcyclist, and tricyclist. (2) Most cars in our traffic scene are very crowded and occluded, which are relatively difficult for detection.

\begin{table}[t]
\renewcommand{\arraystretch}{1.1}
\centering
\footnotesize
\caption{Average precisions (AP) per category on the TJU-DHD-campus.}
\label{tab07}	
\setlength{\tabcolsep}{4.5mm}{
\begin{tabular}{l|c|cc}
\shline
method    & AP   & AP$_{ped}$ & AP$_{rider}$      \\
\shline
(a) RetinaNet \cite{Lin_Focal_ICCV_2017}   &48.4   &50.5 &46.4   \\
(b) FCOS \cite{Tian_FCOS_ICCV_2019}       &49.3   &51.1 &47.5   \\
(c) FPN  \cite{Lin_FPN_CVPR_2017}       &52.4   &54.4 &50.4      \\
(d) Cascade R-CNN \cite{Cai_Cascade_CVPR_2018} &55.1   &57.2 &53.1  \\
\shline
\end{tabular}}
\end{table}

\begin{table}[t]
\renewcommand{\arraystretch}{1.1}
\centering
\footnotesize
\caption{Impact of different input sizes (resolutions) on detection performance on the TJU-DHD-Campus. `$N\times \downarrow\&\uparrow$' indicates first $N$ $\times$ downsampling the images and second  $N$ $\times$ upsampling the images. The run-time (ms) is tested on a single NVIDIA P6000 GPU.}
\label{tab07x}	
\setlength{\tabcolsep}{3.2mm}{
\begin{tabular}{l|c|cccc|c}
\shline
input size  & AP   &  AP$_{t}$ &  AP$_{s}$ & AP$_{m}$  &  AP$_{l}$ &  time\\
\shline
1333$\times$800    & 52.4    & 8.5       & 37.4 & 58.6  & 74.9   & 152   \\
$2\times \downarrow\&\uparrow$    &  50.2  &  6.4    & 32.3 & 56.8  & 74.6  & 152   \\
$4\times \downarrow\&\uparrow$    &   44.0  &   3.3   &   21.1  & 48.7  & 72.1 & 152   \\
\hline
\hline
1333$\times$800    & 52.4    & 8.5       & 37.4 & 58.6  & 74.9   & 152   \\
2000$\times$1200    & 59.8    & 22.2     & 48.6 & 64.4  & 77.4   & 182   \\
2666$\times$1600    & 62.7    & 30.7     & 53.3 & 66.4  & 77.8   & 287   \\
\shline
\end{tabular}}
\end{table}

\subsection{Experiments on the TJU-DHD-campus}
In this subsection, some experiments on the TJU-DHD-campus are conducted. Table \ref{tab06} shows the average precisions (AP) of these four detectors. The anchor-free FCOS is 0.9\% better than the anchor-based RetinaNet. Compared with the one-stage method, the two-stage method performs better. For example, FPN is 4.0\% better and 3.1\% better than FCOS and RetinaNet. Compared with FPN, cascade structure further improves detection performance by 2.7\%. By comparing Table \ref{tab04} and \ref{tab06}, it can be seen that the same detector has a better performance on the TJU-DHD-traffic. The reason can be explained as follows. Compared with TJU-DHD-traffic, TJU-DHD-campus has a rich variance in objects (234,455 \textit{vs} 27,650), seasons (4 \textit{vs} 1), and image resolutions. Thus, TJU-DHD-campus is relatively difficult compared to TJU-DHD-traffic.

Table \ref{tab07} further shows average precisions (AP) on  pedestrian category and rider category. It can be seen that: (1) On TJU-DHD-campus, the detectors also achieve a lower AP on the category of rider. For example, AP of RetinaNet on pedestrian detection is 4.1\% higher than that on rider detection. (2) Cascade structure has a stable improvement on both pedestrian detection and rider detection. For example, Cascade R-CNN is 2.8\% better and 2.7\% better than FPN on pedestrian detection and rider detection, respectively.

To show the advantage of our high-resolution object dataset, we conduct two different experiments based on the detector FPN in Table \ref{tab07x}. One is first downsampling the images and then upsampling them to their original resolutions. We adopt $2\times$ and $4\times$ resampling (denoting as the downsampling/upsampling procedure for simplicity), which leads to 2.2\% and 8.4\% drop in AP, respectively. Moreover, the resolution plays a more important role in small-scale object detection. The AP$_s$ drops by 2.2\% with $2\times$ resampling and 16.3\% with $4\times$ resampling. It demonstrates that high-resolution image quality is useful for improving object detection performance, especially for small-scale object detection performance.

The second experiment is using different input sizes (\textit{i.e.,} 1,333$\times$800 pixels, 2,000$\times$1,200 pixels, and 2,666$\times$1,600 pixels).  With the increment of input size, the detection accuracy becomes higher. When the input size ranges from 1,333$\times$800 pixels to 2,000$\times$1,200 pixels, the AP has 7.4\% improvement. Moreover, the improvement mainly comes from tiny and small objects. When the input size ranges from 1,333$\times$800 pixels to 2,000$\times$1,200 pixels, AP$_t$ and AP$_s$ have 13.7\% and 11.2\% improvements, while AP$_m$ and AP$_l$ have 5.8\% and 2.5\% improvements. When using the high-resolution 2,666$\times$1,600 pixels, it totally has a gain of 10.3\% on AP and a gain of 22.2\% on AP$_t$. It means that higher resolution is very useful to detect small-scale objects. However, detecting objects in higher-resolution images needs much more computational cost. Thus, designing efficient detectors for high-resolution images becomes necessary in the future.

\begin{table}[t]
\renewcommand{\arraystretch}{1.1}
\centering
\footnotesize
\caption{Miss Rates (MR) on the TJU-DHD-Pedestrian. As a reference, miss rate on Citypersons is also shown. In each dataset, results on different subsets are shown.}
\label{tab08}	
\setlength{\tabcolsep}{0.8mm}{
\begin{tabular}{l|c||ccc|cc}
\shline
Method	    &  set	   	          & \textbf{R}    & \textbf{RS}     & \textbf{HO}    & \textbf{R+HO}   & \textbf{A}\\	
\shline
(a) RetinaNet \cite{Lin_Focal_ICCV_2017}	& TJU-Ped-campus		      &34.73	&82.99		 & 71.31	  & 42.26 &44.34\\
(b) FCOS \cite{Tian_FCOS_ICCV_2019}	    & TJU-Ped-campus		      &31.89	&81.28	 & 69.04	  & 39.38	&41.62	\\
(c) FPN	\cite{Lin_FPN_CVPR_2017}        & TJU-Ped-campus		      &27.92	&73.14	 & 67.52	  & 35.67	 	&38.08\\
\hline
(d) RetinaNet \cite{Lin_Focal_ICCV_2017}	& TJU-Ped-traffic		      &23.89	&37.92		 & 61.60	  & 28.45 &41.40	         \\
(e) FCOS \cite{Tian_FCOS_ICCV_2019}	    & TJU-Ped-traffic	 &24.35	&37.40		 & 63.73	  & 28.86	   &40.02     \\
(f) FPN	\cite{Lin_FPN_CVPR_2017}	        & TJU-Ped-traffic	  &22.30	&35.19	 & 60.30	  & 26.71	&37.78	\\
\hline
(g) RetinaNet \cite{Lin_Focal_ICCV_2017}	& Citypersons	      &15.99	&28.54	 & 49.65	  & 32.36	&43.86	 \\
(h) FCOS \cite{Tian_FCOS_ICCV_2019}	    & Citypersons	  &18.29	&27.70	 & 52.42	  & 34.73	&44.39\\
(i) FPN	\cite{Lin_FPN_CVPR_2017}	        & Citypersons	      &14.25	&26.67		 & 49.26	  & 29.71 &38.79\\
\shline
\end{tabular}}
\end{table}

\begin{table}[t]
\renewcommand{\arraystretch}{1.1}
\centering
\footnotesize
\caption{Miss Rates (MR) of FPN by using cross-scene evaluation. TJU-Ped-campus and Ped-traffic in TJU-DHD-Pedestrian, Citypersons, EuroCity are four different datasets.}
\label{tab09}	
\setlength{\tabcolsep}{3.5mm}{
\begin{tabular}{l|c||cc}
\shline
train set     & test set      & \textbf{R}     & \textbf{R+HO}    \\ 	
\shline
\multirow{4}{*}{(a) TJU-Ped-campus}    & TJU-Ped-campus    & 27.92          & 35.67       \\
    & TJU-Ped-traffic    & 30.62         & 33.89    \\
    & Citypersons    & 16.15           & 33.46  \\
    & Eurocity     & 14.10          & 26.73   \\
\cline{2-4}
    &   mean    & 22.20         & 32.43         \\ 
\hline
\hline
\multirow{4}{*}{(b) TJU-Ped-traffic}    & TJU-Ped-campus     & 42.08          & 50.55     \\
    & TJU-Ped-traffic    & 22.30           & 26.71 \\
    & Citypersons    & 23.79           & 43.09     \\
    & Eurocity    &   27.37        & 44.14  \\
\cline{2-4}
   & mean      & 28.89         & 41.12         \\ 
\hline
\hline
\multirow{4}{*}{(c) Citypersons}    & TJU-Ped-campus     & 47.70           & 54.77   \\
    & TJU-Ped-traffic    & 46.27          & 49.40  \\
    & Citypersons    & 14.25           & 29.71    \\
    & Eurocity    & 21.99      & 35.58  \\
\cline{2-4}
     &   mean   & 32.55    & 42.37         \\ 
\hline
\hline
\multirow{4}{*}{(d) Eurocity}   & TJU-Ped-campus & 41.86     & 48.52 \\
    & TJU-Ped-traffic   &    43.27       & 46.12 \\
    & Citypersons    &    16.16       & 31.51    \\
    & Eurocity    & 6.81          & 15.83 \\
\cline{2-4}
     &   mean   & 27.78    & 35.50        \\      
\shline
\end{tabular}}
\end{table}

\begin{table*}[t]
\renewcommand{\arraystretch}{1.1}
\centering
\footnotesize
\caption{Miss Rates (MR) on the Citypersons by fine-tuning FPN on our TJU-DHD-pedestrian.}
\label{tab10}	
\setlength{\tabcolsep}{5.9mm}{
\begin{tabular}{l|ccc|cc}
\shline
Method	  & \textbf{R}    & \textbf{RS}     & \textbf{HO}    & \textbf{R+HO}   & \textbf{A}\\ 
\shline
(a) Citypersons  \cite{Zhang_CityPersions_CVPR_2017}       & 14.25      & 26.67          & 49.26        & 29.71     & 38.79\\
(b) only TJU-Ped-traffic $\rightarrow$	Citypersons          & 13.23      & 22.81          & 48.93        & 29.01      & 37.38\\
(c) only TJU-Ped-campus $\rightarrow$	Citypersons          & 10.68      & 18.99           & 41.15        & 24.40    & 32.44    \\
(d) TJU-DHD-pedestrian $\rightarrow$	Citypersons         & 10.13      & 17.07         & 40.40        & 23.88       & 31.89   \\
\shline
\end{tabular}}
\end{table*}

\begin{figure*}[t]
	\centering
	\includegraphics[width=2.35in]{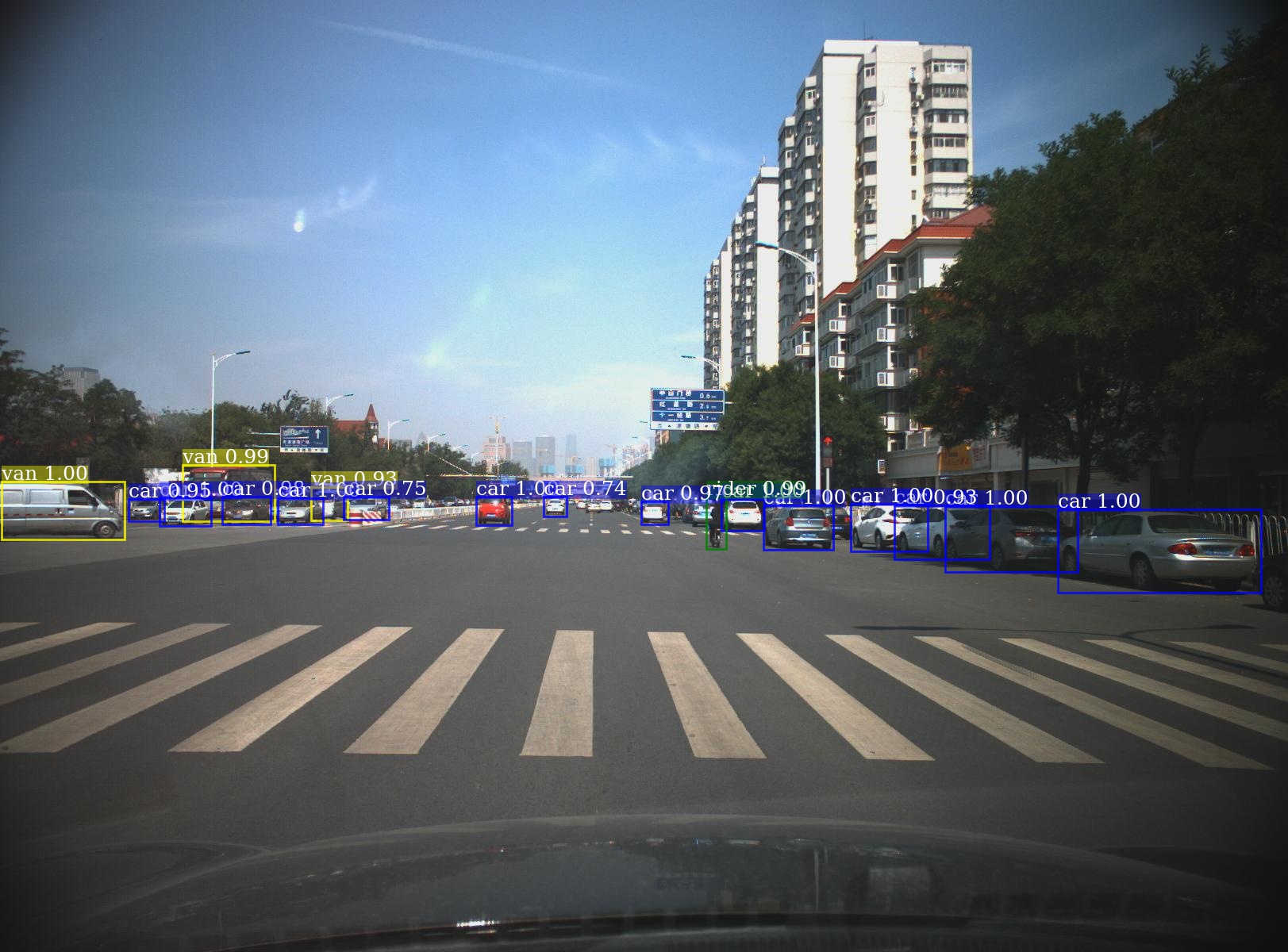}
	\includegraphics[width=2.35in]{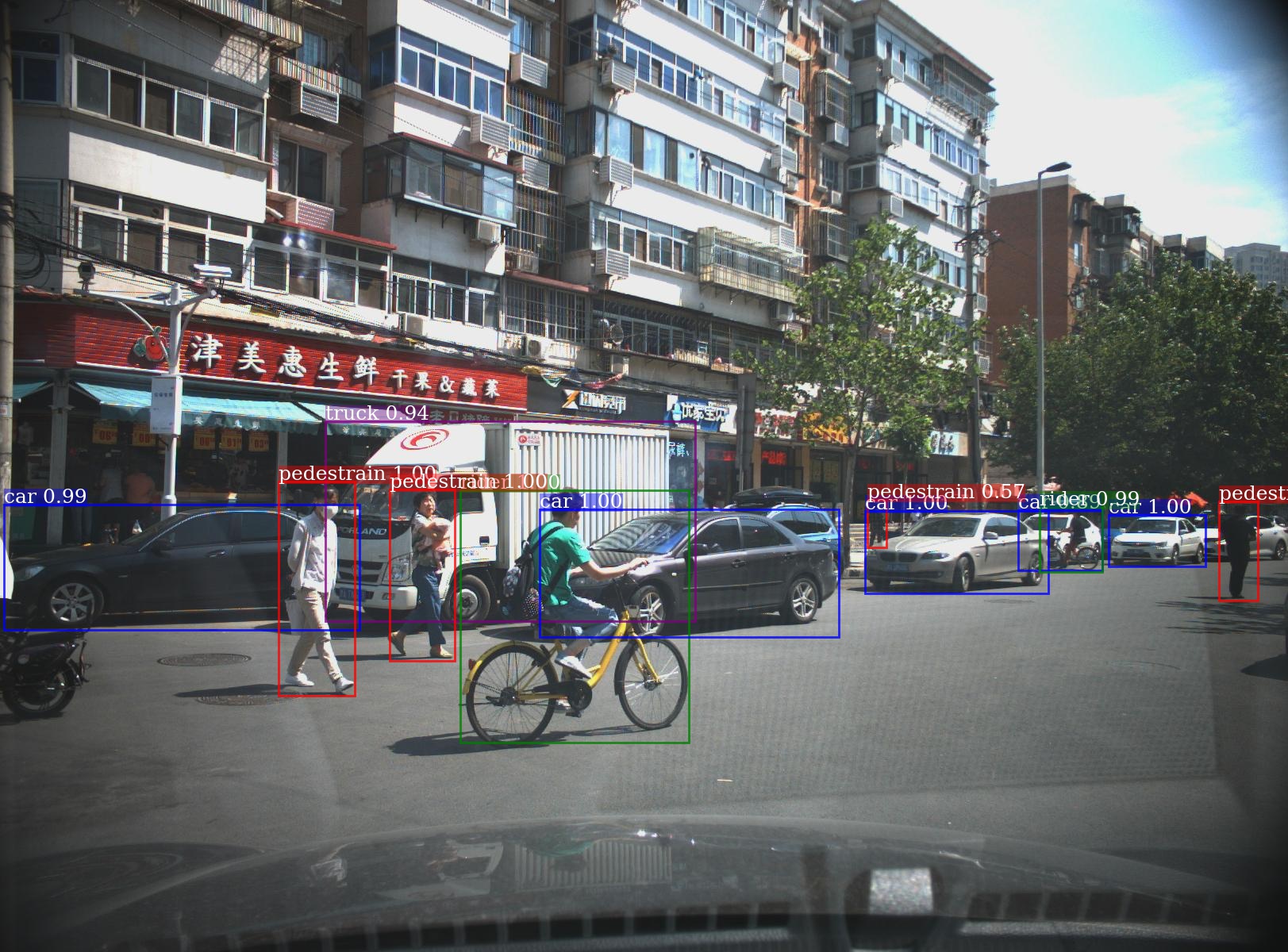}
	\includegraphics[width=2.35in]{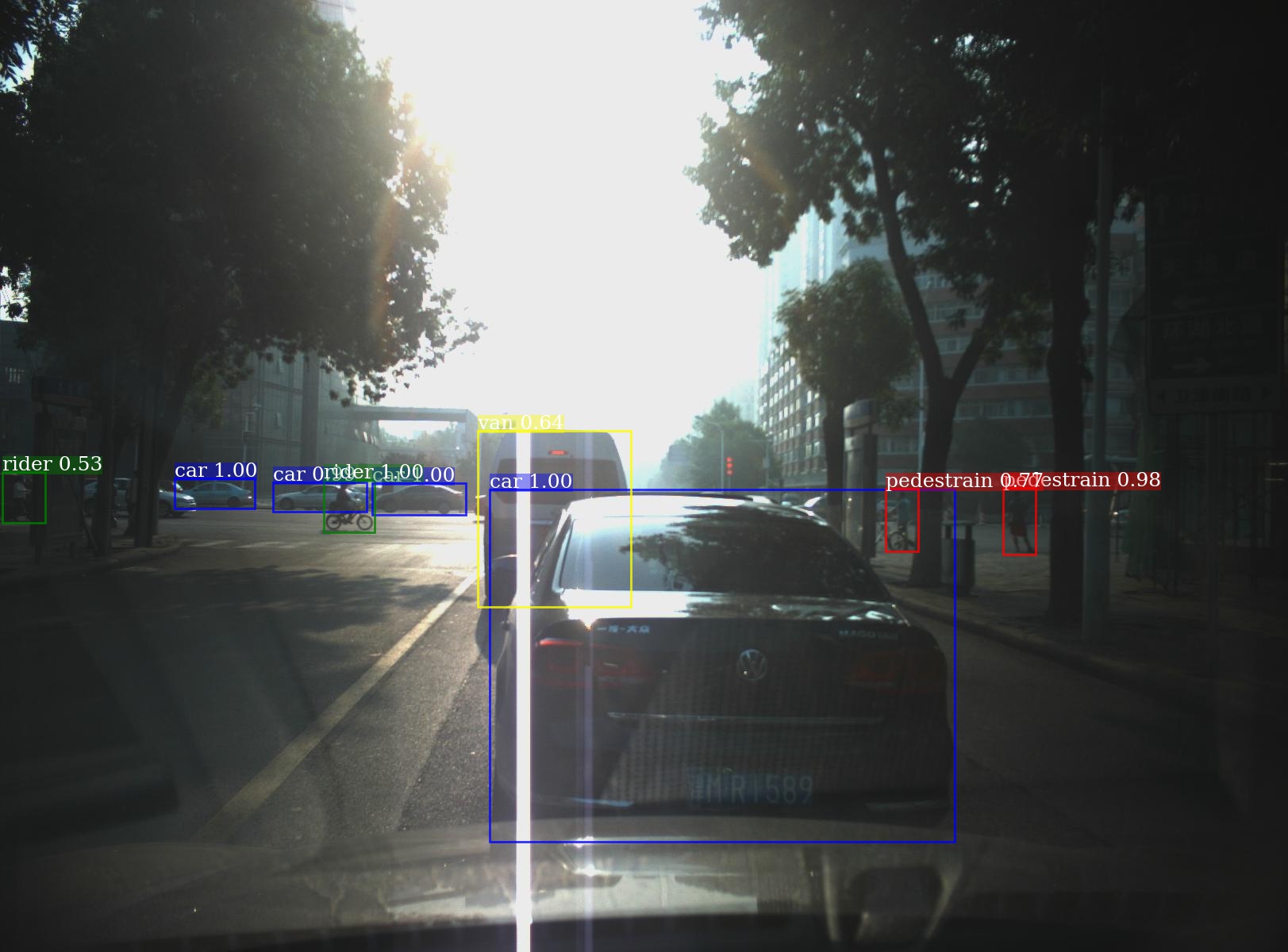}
	\includegraphics[width=2.35in]{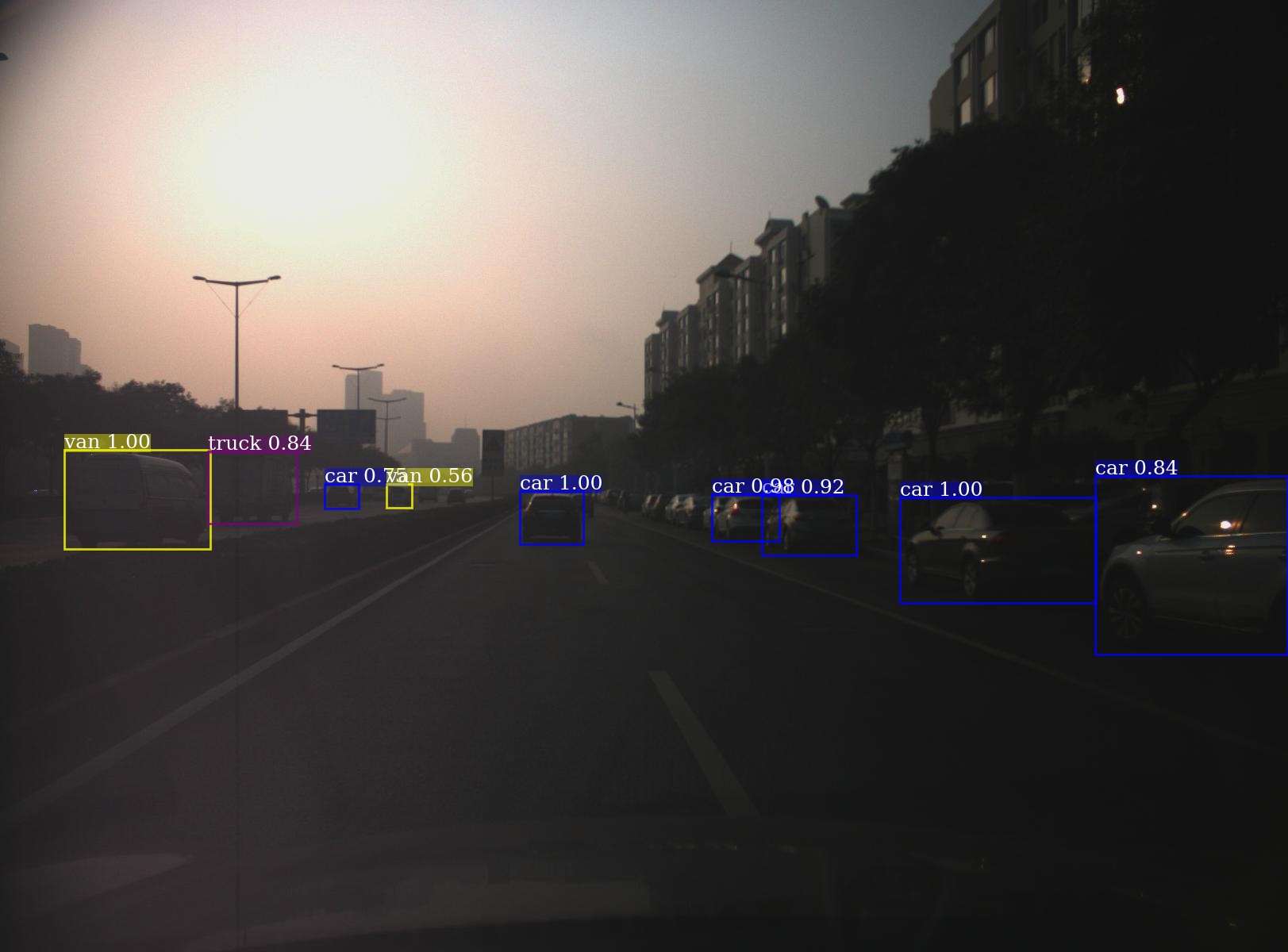}
	\includegraphics[width=2.35in]{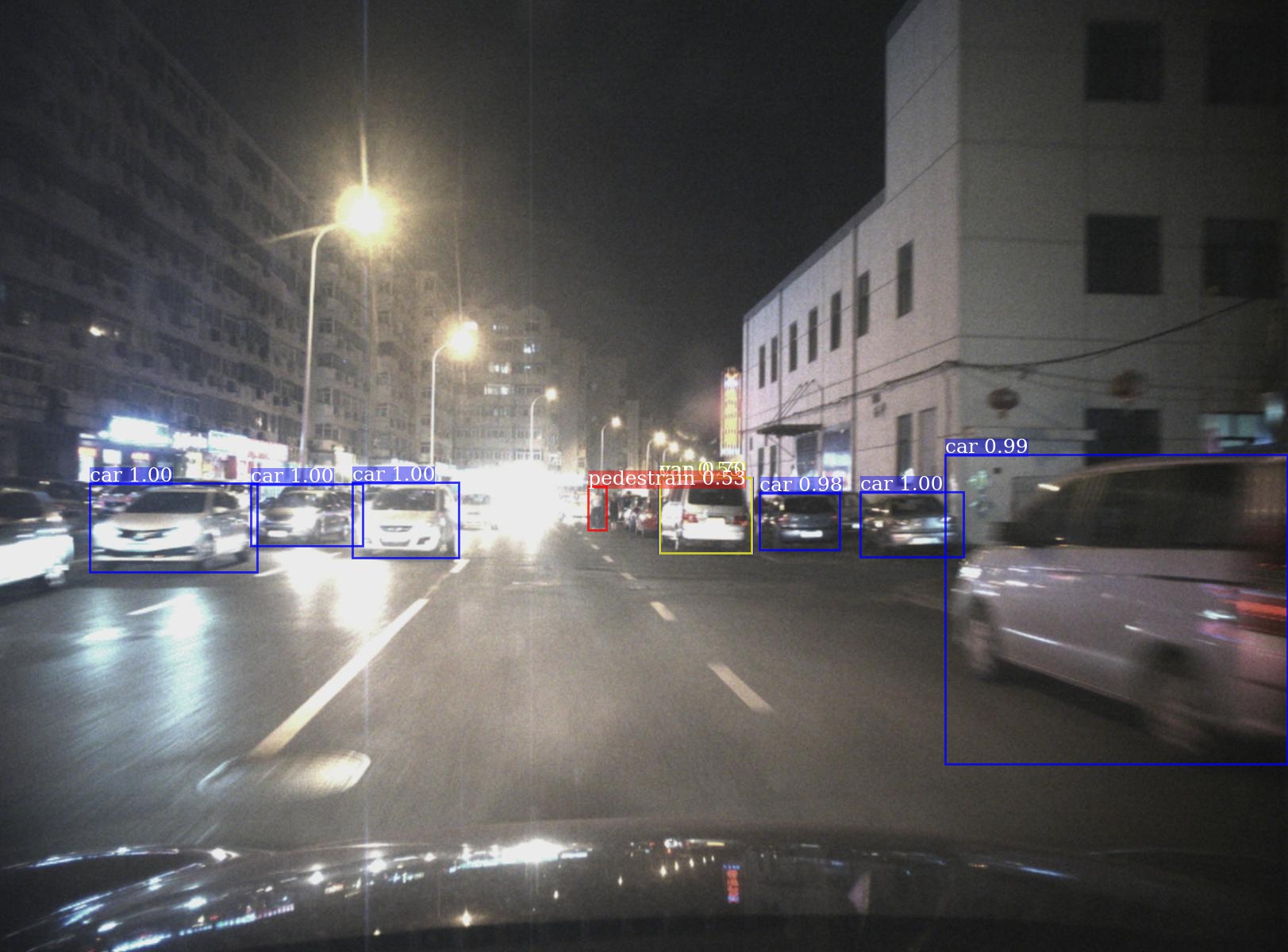}
	\includegraphics[width=2.35in]{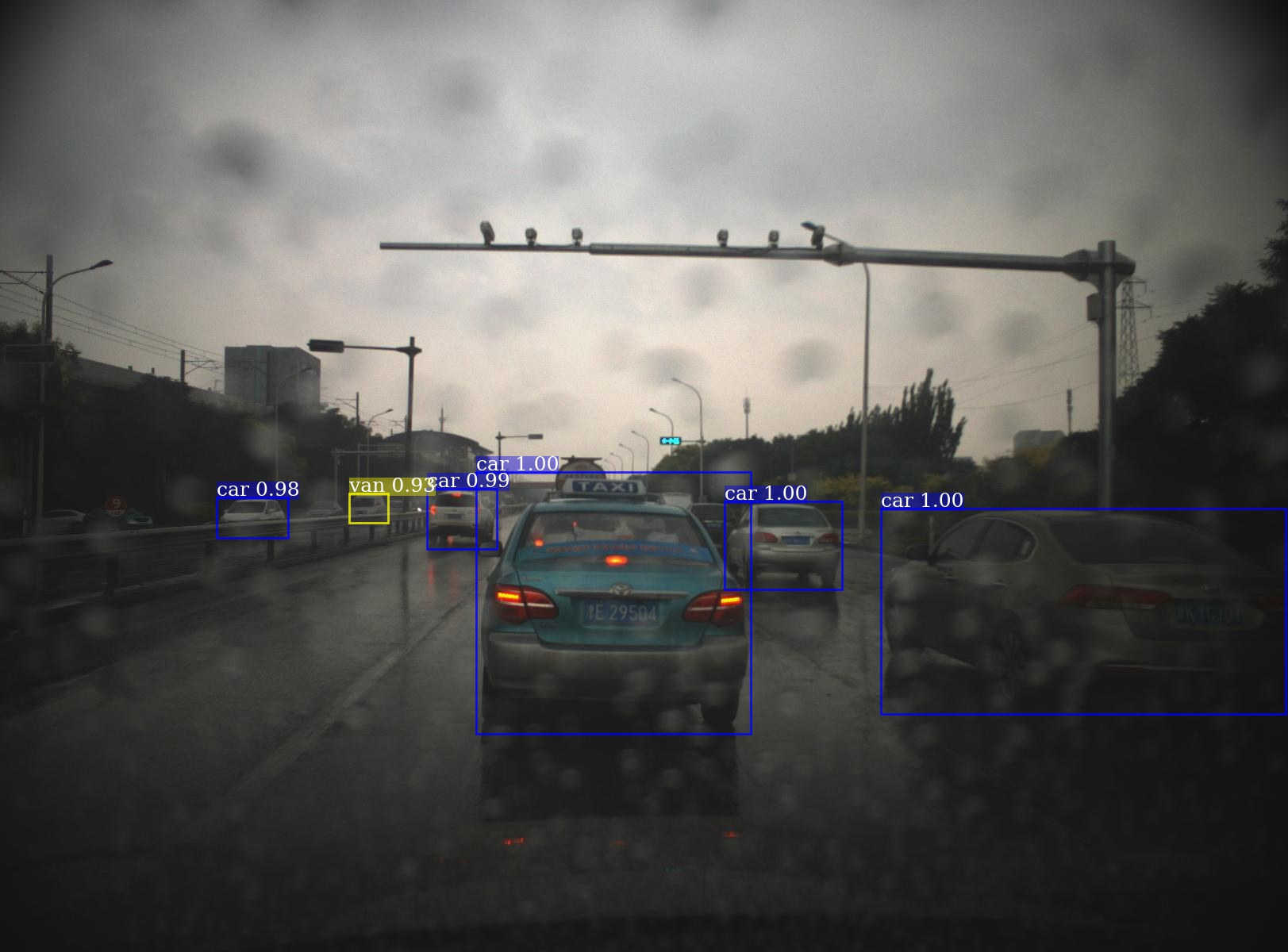}
	\caption{Qualitative results of Cascade R-CNN on the TJU-DHD-traffic. The images under different illuminations and different weathers are chosen.}
	\label{fig07}
\end{figure*}
\begin{figure*}[t]
	\centering
	\includegraphics[width=2.35in]{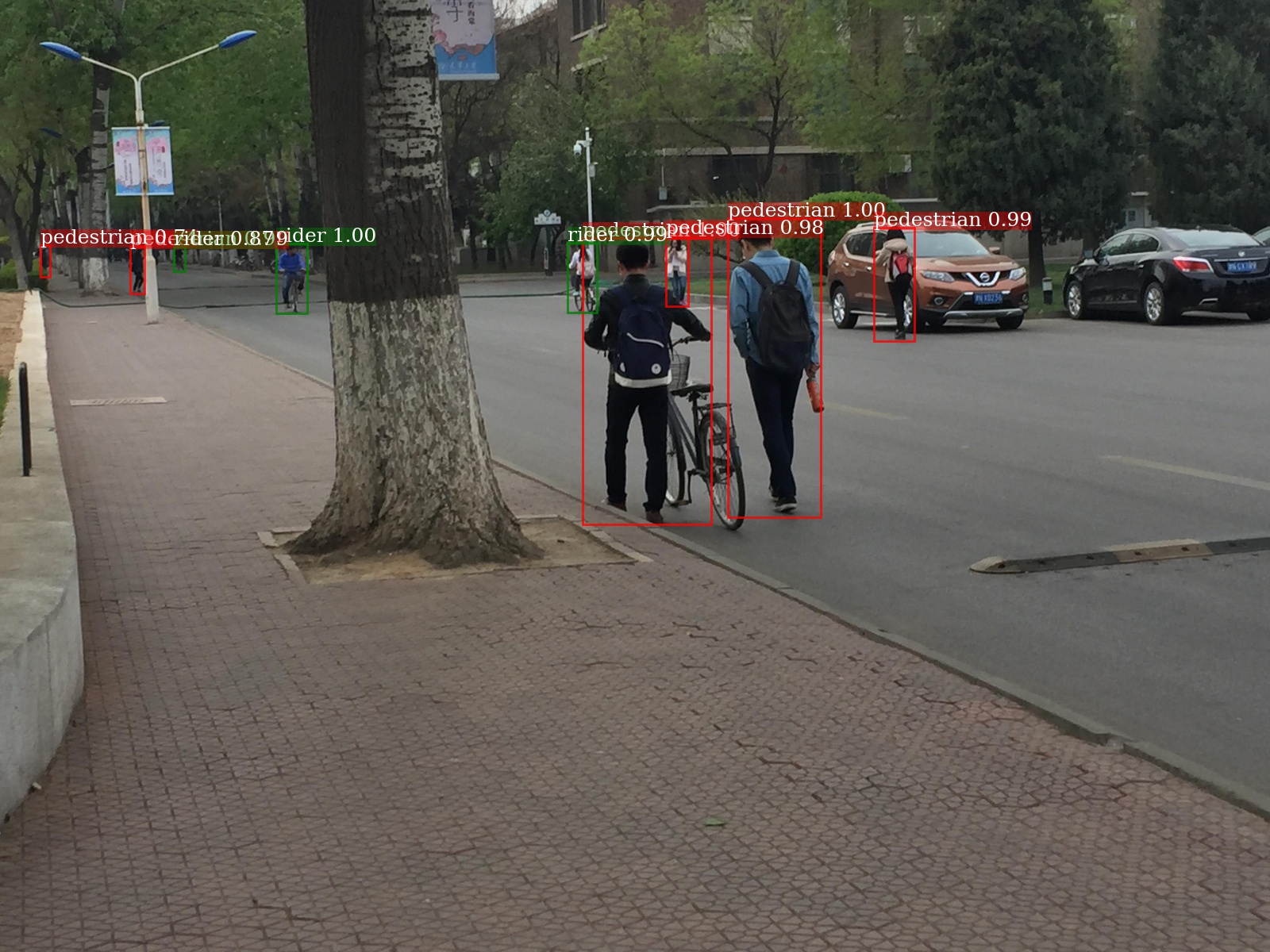}
	\includegraphics[width=2.35in]{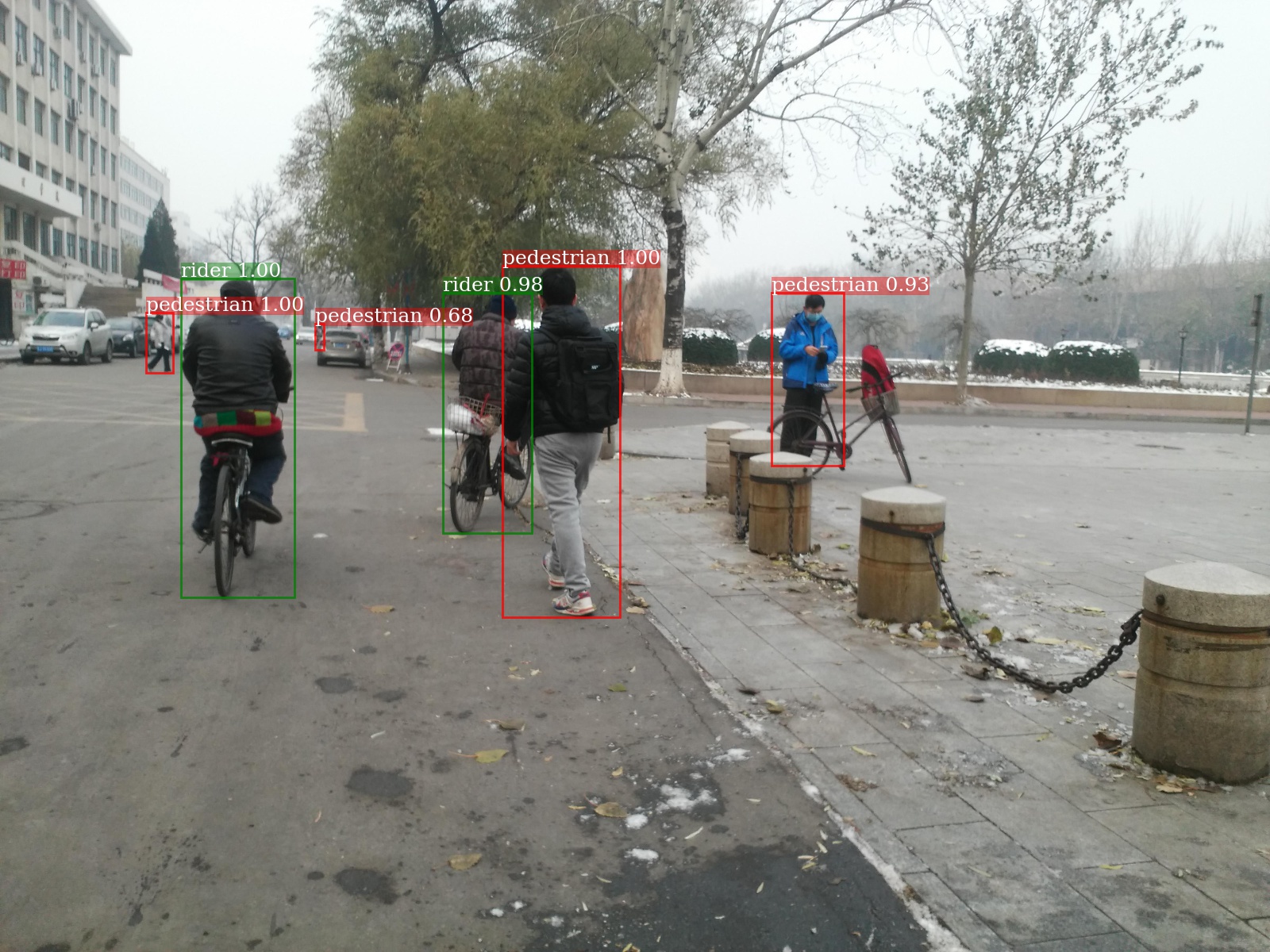}
	\includegraphics[width=2.35in]{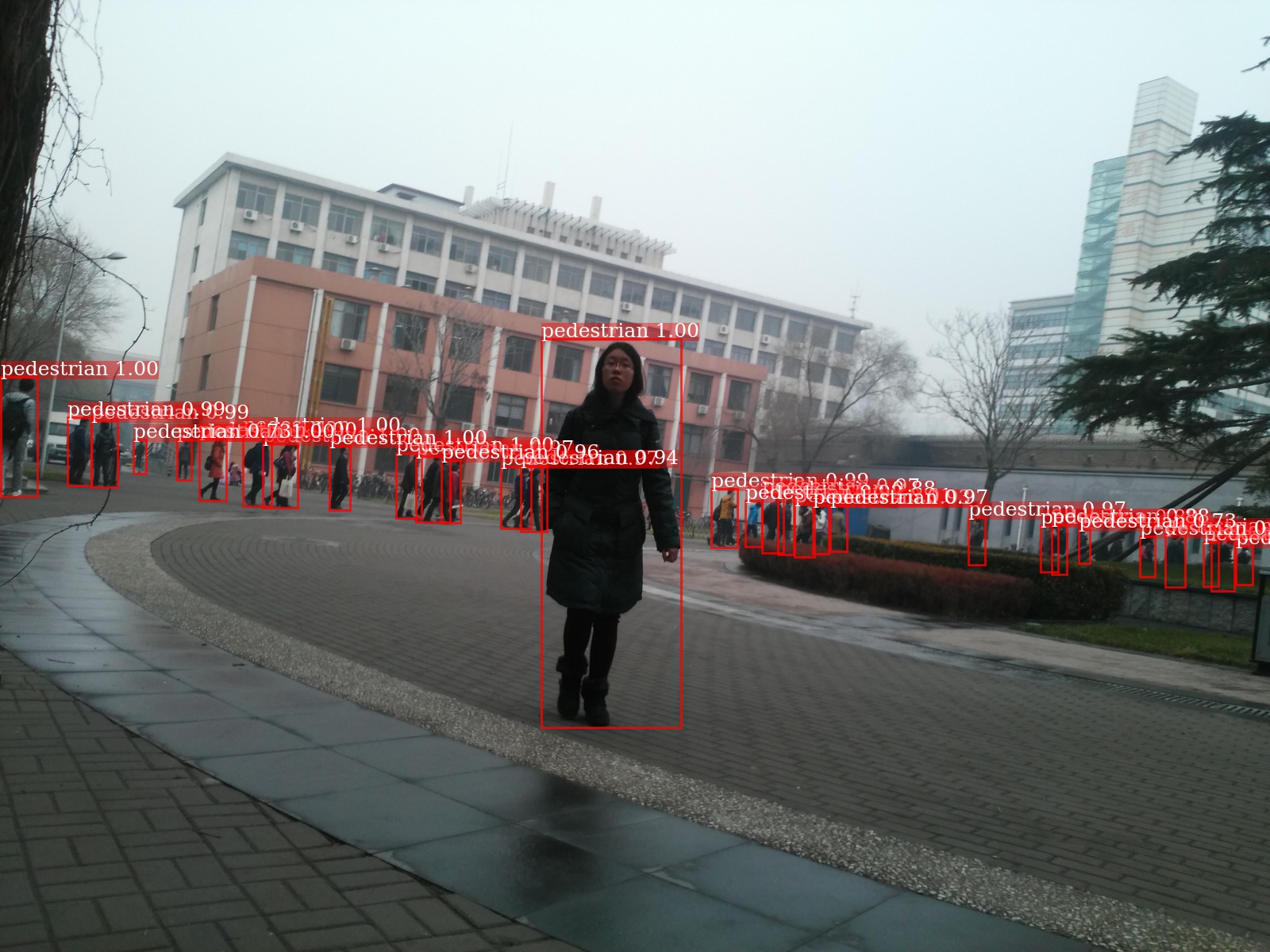}
	\includegraphics[width=2.35in]{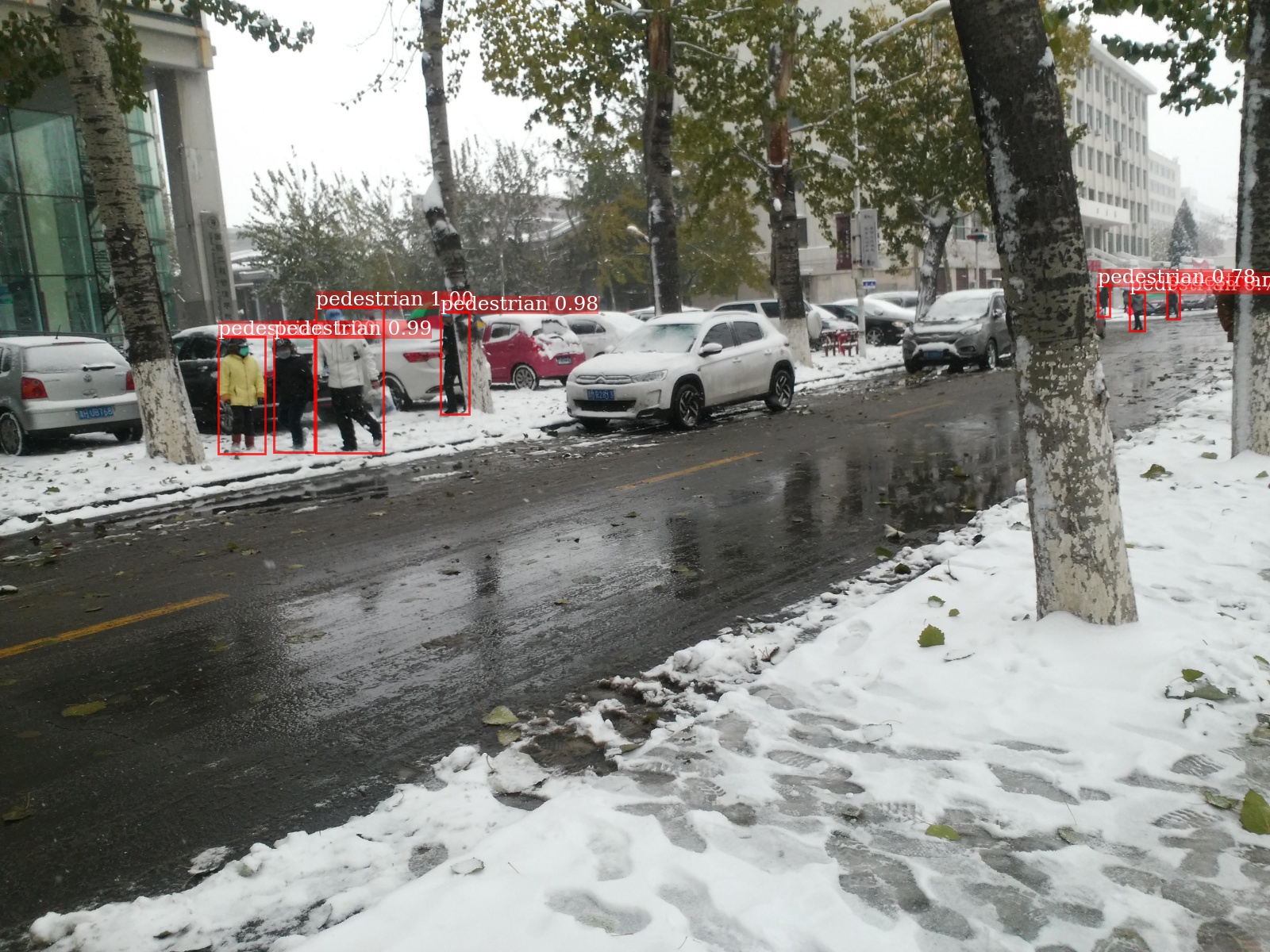}
	\includegraphics[width=2.35in]{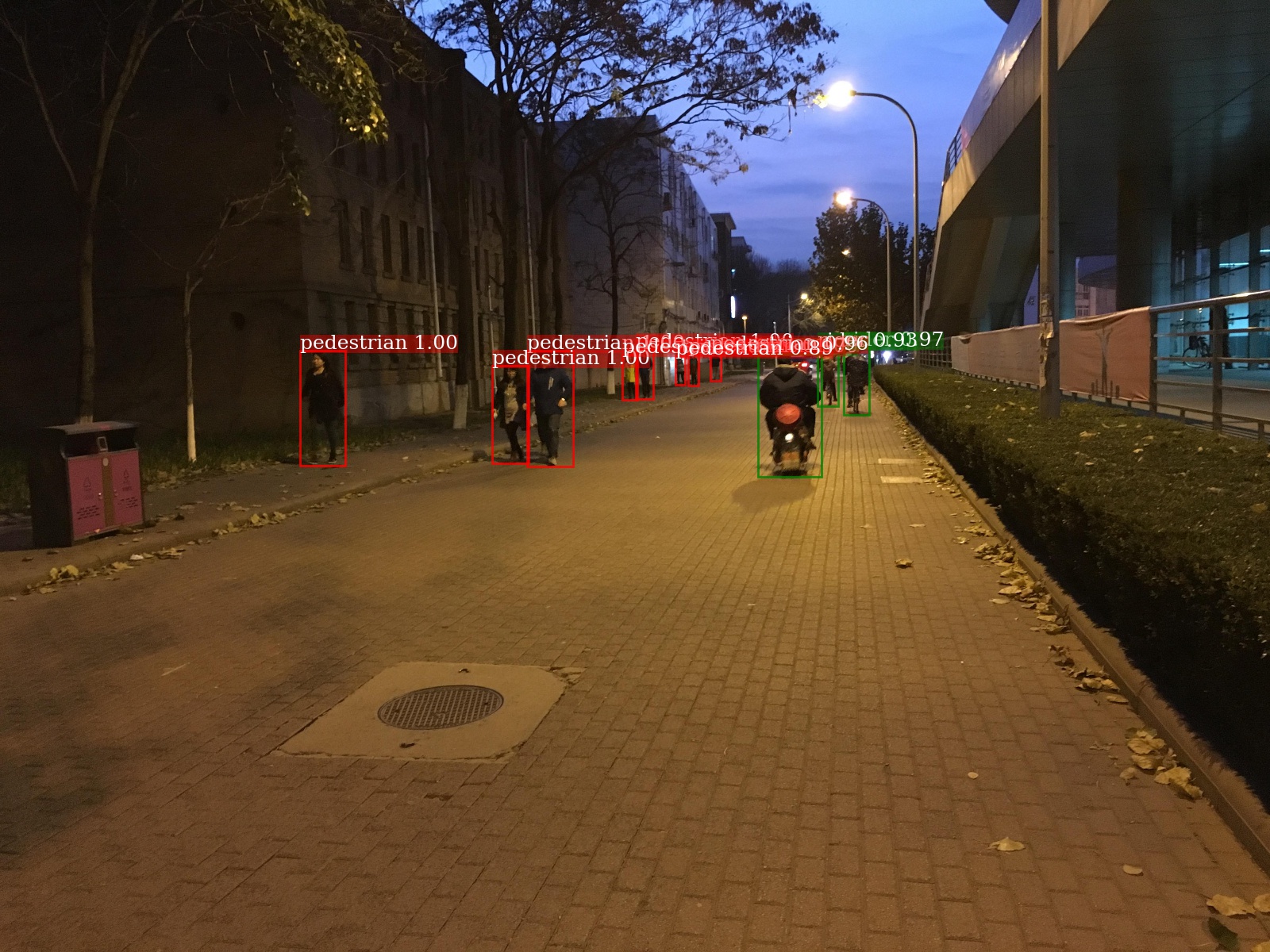}
	\includegraphics[width=2.35in]{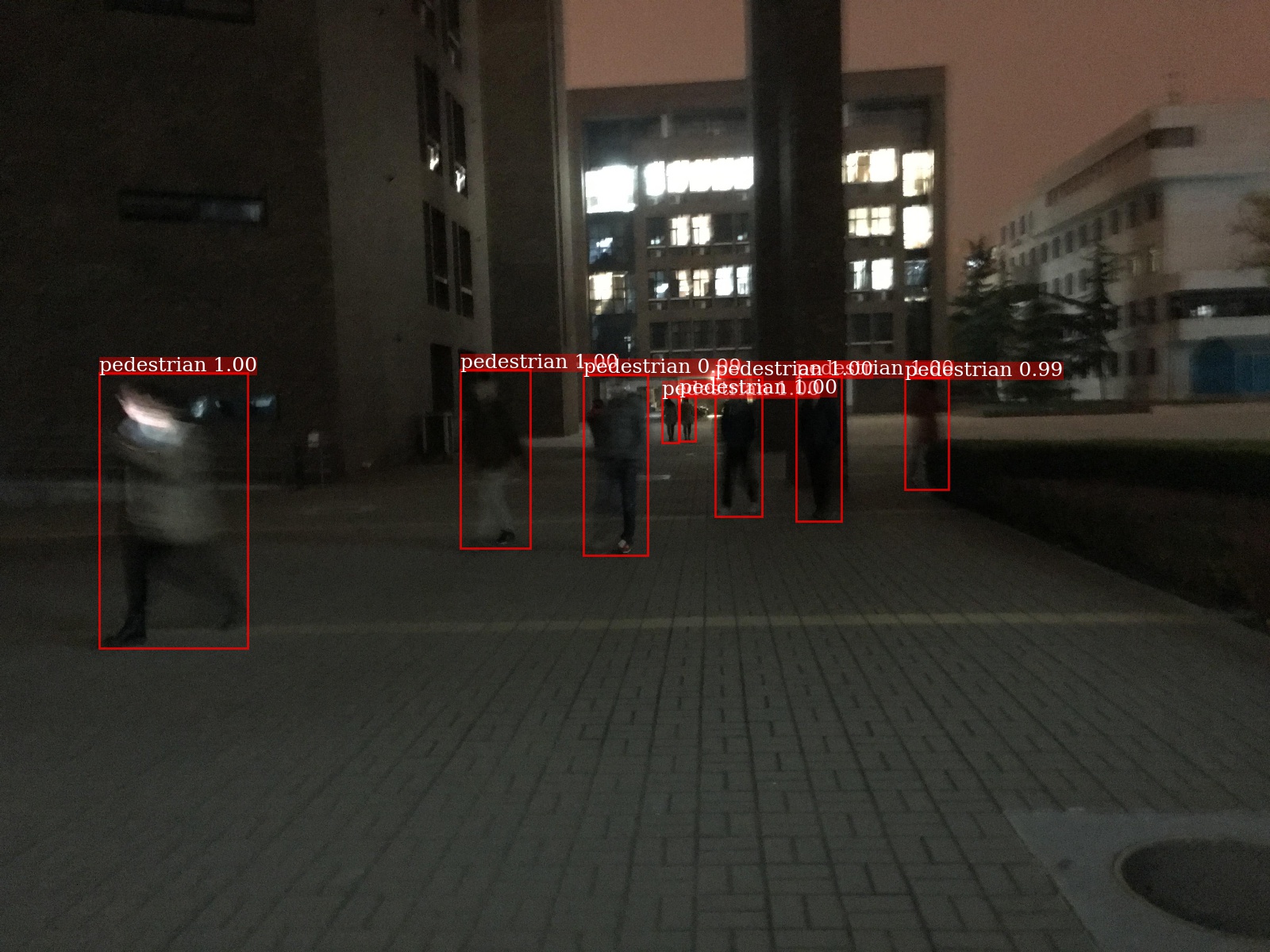}
	\caption{Qualitative results of Cascade R-CNN on TJU-DHD-campus. The images under different illuminations and different seasons are chosen.}
	\label{fig08}
\end{figure*}

\begin{figure*}[!t]
	\centering
	\includegraphics[width=7.2in]{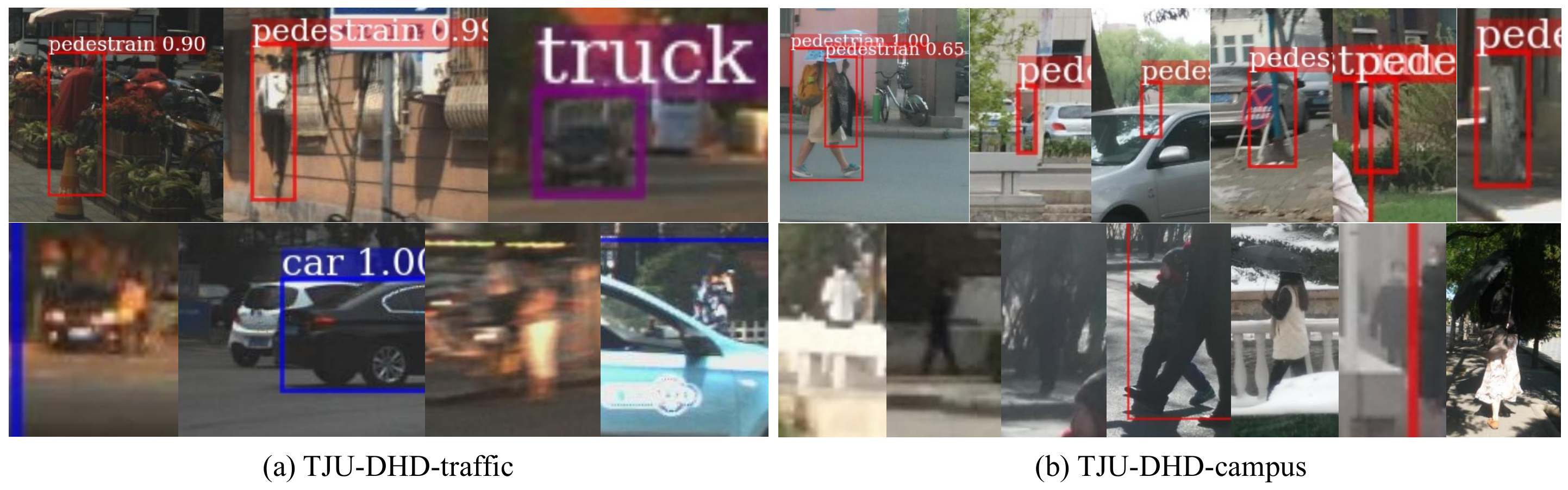}
	\caption{Some failure cases on the TJU-DHD-traffic and TJU-DHD-campus. The top row shows the false objects, and the bottom row shows the missed objects.}
	\label{fig09}
\end{figure*}

\subsection{Experiments on the TJU-DHD-pedestrian}
In this subsection, some experiments based on RetinaNet \cite{Lin_Focal_ICCV_2017}, FCOS \cite{Tian_FCOS_ICCV_2019}, and FPN \cite{Lin_FPN_CVPR_2017} are conducted on the TJU-DHD-pedestrian. Similar to the Citypersons dataset \cite{Zhang_CityPersions_CVPR_2017}, miss rates under the different subsets are shown. These subsets are the reasonable set (\textbf{R}), the reasonable small set (\textbf{RS}), the heavy occlusion set (\textbf{HO}), the reasonable set and the heavy occlusion set (\textbf{R+HO}), and the all set (\textbf{A}).

Table \ref{tab08} shows miss rates of the same-scene evaluation on the TJU-DHD-pedestrian. The detector is trained and tested on the same scene dataset (\textit{i.e.,} either TJU-Ped-campus or TJU-Ped-traffic). As a reference, miss rates on the Citypersons are also shown. It can be seen as follows: (1) Similar to object detection, the two-stage detector FPN outperforms the single-stage detector RetinaNet and the anchor-free detector FCOS on pedestrian detection. For example, FPN outperforms RetinaNet and FCOS by 6.26\% and 3.54\% on the \textbf{A} set of 
TJU-Ped-campus. (2) TJU-Ped-campus and TJU-Ped-traffic are more challenging than Citypersons. For example, miss rates of FPN are 27.92\% and 22.30\% on the \textbf{R} set of TJU-Ped-campus and TJU-Ped-traffic, while the miss rate of FPN is 14.25\% on the \textbf{R} set of Citypersons. Namely, miss rates of FPN on the \textbf{R} set of TJU-Ped-campus and TJU-Ped-traffic are 13.67\% and 8.05\% higher than that on the \textbf{R} set of Citypersons. Thus, there is a large space to improve the performance on our newly built pedestrian dataset.

To show the diversity of the built TJU-DHD-pedestrian quantitatively, Table \ref{tab09} further gives a cross-scene evaluation on the TJU-Ped-campus, TJU-Ped-traffic, Citypersons \cite{Zhang_CityPersions_CVPR_2017}, and EuroCity persons \cite{Braun_EuroCity_PAMI_2019} based on two-stage FPN. The miss rate on each dataset and the mean miss rate over the four datasets are both given. If one dataset has a better diversity than another dataset, we think that it will have a lower mean miss rate when performing the cross-scene evaluation.  It can be seen as follows: (1) The detector trained on TJU-Ped-campus has a best diversity, which achieves the lowest mean miss rate. For example, the detector trained on TJU-Ped-campus achieves 22.20\% mean miss rate on the \textbf{R}, while that trained on Citypersons achieves 32.55\%  mean miss rate on the \textbf{R}. Namely, the detector trained on TJU-Ped-campus outperforms that trained on Citypersons by 10.35\%. Similarly, TJU-Ped-campus outperforms EuroCity by 5.58\%. It is demonstrated that our TJU-DHD-pedestrian has a richer diversity compared to EuroCity. (2) Our TJU-Ped-campus achieves a lower mean miss rate than our TJU-Ped-traffic. The reason can be explained as follows: Compared with TJU-Ped-traffic, TJU-Ped-campus has a rich variance in objects (234,455 \textit{vs} 27,650) and seasons (4 \textit{vs}  1).  (3) Though TJU-Ped-traffic and EuroCity are collected from the traffic scenes, they have a large domain gap. For example, the detectors trained on these two datasets have 20.97\% difference when testing on TJU-Ped-traffic, 20.56\% difference when testing on EuroCity. It means that the traffic scenes in China and Europe have a large difference. Thus, these two datasets are complementary.

Finally, we fine-tune the two-stage detector FPN on the recently built Citypersons by using our TJU-DHD-pedestrian in Table \ref{tab10}. Namely, the detector is firstly trained based on our TJU-DHD-pedestrian and secondly fine-tuned on the Citypersons dataset. The initial miss rate on the \textbf{R} set of Citypersons is 14.25\%. If only using TJU-Ped-traffic, miss rate on the \textbf{R} set drops from 14.25\% to 13.23\%. If only using TJU-Ped-campus, miss rate on the \textbf{R} set drops from 14.25\% to 10.68\%. If using both TJU-Ped-traffic and TJU-Ped-campus, miss rate on the \textbf{R} set drops from 14.25\% to 10.13\%. The total improvement using our TJU-DHD-pedestrian is 4.12\% on the \textbf{R} set. Namely, our TJU-DHD-pedestrian can help improve the detection performance on the Citypersons dataset.

\subsection{Visualizations}
In this subsection, some visualizations on both TJU-DHD-traffic and TJU-DHD-campus are given. Because pedestrian detection is a special case of object detection in our dataset, the visualizations of pedestrian detection results on the TJU-DHD-pedestrian are not further given. Fig. \ref{fig07} and Fig. \ref{fig08} respectively show detection results of Cascade R-CNN \cite{Cai_Cascade_CVPR_2018} on the TJU-DHD-traffic and TJU-DHD-campus. To better see the performance of Cascade R-CNN on different conditions, the images under different illumination variance, different weathers, and different seasons are chosen for visualizations. It can be seen that Cascade R-CNN achieves a good result under these variances in some degree.

Meanwhile, some failure cases, including the false positives and the false negatives, of Cascade R-CNN on the TJU-DHD-traffic and TJU-DHD-campus are further given in Fig. \ref{fig09}. The top row shows examples of the false objects. It can be seen that some object-like things (\textit{e.g.,} air-conditioners and street signs), object-like background (\textit{e.g.,} potted plants), and some parts of objects are easily recognized as the detected objects. The bottom row gives examples of missed objects. It can be seen that the occluded objects, the small-scale objects, and low-illumination objects are usually missed by the detector. These problems remain unsolved and might be the key for further research in object detection.

\subsection{Discussion}
\textit{Challenges} Based on the above qualitative and quantitative experiments, we give a summary about the main challenges of object detection and pedestrian detection as follows: (1) Small-scale object detection and occluded object detection are still the bottlenecks of object detection. On the one hand, small-scale objects and occluded objects both have limited useful information. Small-scale objects have a noisy and blurred appearance and contain limited useful information, while occluded objects lose some local part information. On the other hand, small-scale objects and occluded objects make the intra-class variance large. (2) Most of the object detectors have a poor generality ability. For example, the detector trained on EuroCity persons has a poor performance on our TJU-Ped-campus (see Table \ref{tab09}). However, to meet the actual application requirements, the object detector needs to have good robustness for domain adaption.

\textit{Future Directions} Based on the challenges mentioned above, some future directions are summarized as follows: (1) \textit{Small-scale object detection}. One direction is that designing better feature pyramid and image pyramid structures. Another direction is that designing efficient detectors for high-resolution object detection. As a result, it does not add much computational cost when using high-resolution images. (2) \textit{Occluded object detection}. One direction is that making full use of visible regions to suppress the negative impact of occluded regions. Another direction is that exploiting how to retain correct detection in the crowded object detection.  (3) \textit{Cross-scene object detection}. Traffic scene and campus scene are two common scenes, which have important application value. How to make the detector have a good generation ability is an important topic in the future.

\section{Conclusion}
In this paper, a new diverse high-resolution dataset for object detection in two typical scenes (traffic scene and campus scene) was built. The experiments based on four typical detectors are conducted to give the baseline performance on this dataset. Meanwhile, based on this dataset, a new large-scale pedestrian dataset is also built.  The built datasets, TJU-DHD-traffic, TJU-DHD-campus, and TJU-DHD-pedestrian, can hopefully be one of the fundamental benchmarks for object detection and pedestrian detection.


\begin{thebibliography}{1}
\bibitem{Braun_EuroCity_PAMI_2019}
M. Braun, S. Krebs, F. Flohr, and D. M. Gavrila,
``The EuroCity persons dataset: A novel benchmark for object detection,''
\newblock {\em IEEE Trans. Pattern Analysis and Machine Intelligence}, vol. 41, no. 8, pp. 1844-1861, 2019.

\bibitem{Brazil_SDS_ICCV_2017}
G. Brazil, X. Yin, and X. Liu,
\newblock ``Illuminating pedestrians via simultaneous detection and segmentation,''
\newblock {\em Proc. IEEE International Conference on Computer Vision}, 2017.


\bibitem{Brazil_ANP_CVPR_2019}
G. Brazil and X. Liu,
\newblock ``Pedestrian detection With autoregressive network phases,''
\newblock {\em Proc. IEEE Conference on Computer Vision and Pattern	Recognition}, 2019.



\bibitem{Cai_MSCNN_ECCV_2016}
Z.~Cai, Q.~Fan, R.~S. Feris, and N.~Vasconcelos,
\newblock ``A unified multi-scale deep convolutional neural network for fast
    object detection,''
\newblock {\em Proc. European Conference on Computer Vision}, 2016.

\bibitem{Cai_Cascade_CVPR_2018}
Z.~Cai and N.~Vasconcelos,
\newblock ``Cascade r-cnn: Delving into high quality object detection,''
\newblock {\em Proc. IEEE Conference on Computer Vision and Pattern	Recognition}, 2018.

\bibitem{Cao_TripleNet_CVPR_2019}
J. Cao, Y. Pang, and X. Li,
\newblock ``Triply supervised decoder networks for joint detection and segmentation,''
\newblock {\em Proc. IEEE Conference on Computer Vision and Pattern	Recognition}, 2019.


\bibitem{Cao_NNNF_CVPR_2016}
J. Cao, Y. Pang, and X. Li,
\newblock ``Pedestrian detection inspired by appearance constancy and shape symmetry,''
\newblock {\em Proc. IEEE Conference on Computer Vision and Pattern Recognition}, 2016.

\bibitem{Cao_MCF_TIP_2016}
J. Cao, Y. Pang, and X. Li,
\newblock ``Learning multilayer channel features for pedestrian detection,''
\newblock {\em IEEE Transactions on Image Processing}, 2016.


\bibitem{Cao_HSD_ICCV_2019}
J. Cao, Y. Pang, J. Han, and X. Li,
\newblock ``Hierarchical shot detector,''
\newblock {\em Proc. IEEE International Conference on Computer Vision}, 2019.


\bibitem{Cao_MHN_TCSVT_2019}
J. Cao, Y. Pang, S. Zhao, and X. Li,
\newblock ``High-level semantic networks for multi-scale object detection,''
\newblock {\em IEEE Trans. on Circuits and Systems for Video Technology }, 2019.


\bibitem{Chi_PedHunter_AAAI_2020}
C. Chi, S. Zhang, J. Xing, Z. Lei, S. Z. Li, and X. Zou,
\newblock ``PedHunter: Occlusion robust pedestrian detector in crowded scenes,''
\newblock {\em Proc. AAAI Conference on Artificial Intelligence}, 2020.

\bibitem{Cordts_Cityscapes_CVPR_2016}
M. Cordts, M. Omran, S. Ramos, T. Rehfeld, M. Enzweiler, R. Benenson, U. Franke, S. Roth, and B. Schiele, 
``The Cityscapes dataset for semantic urban scene understanding,''
\newblock {\em Proc. IEEE Conference on Computer Vision and Pattern Recognition}, 2016.


\bibitem{Dalal_HOG_CVPR_2005}
N. Dalal and B. Triggs, ``Histograms of oriented gradients for
human detection,'' 
\newblock {\em Proc. IEEE Conference on Computer Vision and Pattern Recognition}, 2005.


\bibitem{Dollar_ICF_BMVC_2009}
P.~Doll{\'a}r, Z.~Tu, P.~Perona, and S.~Belongie,
\newblock ``Integral channel features,"
\newblock {\em Proc. British Machine Vision Conference}, 2009.

\bibitem{Dollar_PD_PAMI_2012}
P.~Doll{\'a}r, C.~Wojek, B.~Schiele, and P.~Perona, ``Pedestrian detection: An evaluation of the state of the art,''
\newblock {\em IEEE Trans. Pattern Analysis and Machine Intelligence}, vol. 34, no. 4, pp. 743-761, 2012.

\bibitem{Ess_ETH_ICCV_2007}
A. Ess, B. Leibe, and L. Van Gool, 
``Depth and appearance for mobile scene analysis,'' 
\newblock {\em Proc. IEEE International Conference on Computer Vision}, 2007.

\bibitem{Enzweiler_Daimler_TPAMI_2009}
M. Enzweiler and D. M. Gavrila, 
``Monocular pedestrian detection: Survey and experiments,'' 
\newblock {\em IEEE Trans. Pattern Analysis and Machine Intelligence}, vol. 31, no. 12, pp. 2179-2195, 2009


\bibitem{Everingham_VOC_IJCV_2010}
M.~Everingham, L.~V. Gool, C.~K. Williams, J.~Winn, and A.~Zisserman,
\newblock ``The pascal visual object classes (voc) challenge,''
\newblock {\em Intetnational Journal of Computer Vision}, vol. 88, no. 2, pp. 303--338, 2010.


\bibitem{Duan_CenterNet_ICCV_2019}
K. Duan, S. Bai, L. Xie, H. Qi, Q. Huang, and Q. Tian,
\newblock ``CenterNet: Keypoint triplets for object detection,''
\newblock {\em Proc. IEEE International Conference on Computer Vision}, 2019.


\bibitem{Geiger_KITTI_CVPR_2012}
A.~Geiger, P.~Lenz, and R.~Urtasun, ``Are we ready for autonomous driving? the kitti vision benchmark suite,''
\newblock {\em Proc. IEEE Conference on Computer Vision and Pattern Recognition}, 2012.

\bibitem{Girshick_RCNN_CVPR_2014}
R.~Girshick, J.~Donahue, T.~Darrell, and J.~Malik,
\newblock ``Rich feature hierarchies for accurate object detection and semantic
segmentation,''
\newblock {\em Proc. IEEE Conference on Computer Vision and Pattern
	Recognition}, 2014.

\bibitem{Girshick_FastRCNN_ICCV_2015}
R.~Girshick,
\newblock ``Fast r-cnn,''
\newblock {\em Proc. IEEE International Conference on Computer Vision}, 2015.

\bibitem{Gupta_LVIS_CVPR_2019}
A. Gupta, P. Dollar, and R. Girshick,
``LVIS: A dataset for large vocabulary instance segmentation,''
\newblock {\em Proc. IEEE Conference on Computer Vision and Pattern Recognition}, 2019.

\bibitem{Redmon_YOLO_CVPR_2016}
J.~Redmon, S.~Divvala, R.~Girshick, and A.~Farhadi,
\newblock ``You only look once: Unified, real-time object detection,''
\newblock {\em Proc. IEEE Conference on Computer Vision and Pattern
	Recognition}, 2016.

\bibitem{Ren_FasterRCNN_NIPS_2015}
S.~Ren, K.~He, R.~Girshick, and J.~Sun,
\newblock ``Faster r-cnn: Towards real-time object detection with region proposal networks,''
\newblock {\em Proc. Advances in Neural Information Processing Systems}, 2015.



\bibitem{Kong_RON_CVPR_2017}
T.~Kong, F.~Sun, A.~Yao, H.~Liu, M.~Lu, and Y.~Chen,
\newblock ``Ron: Reverse connection with objectness prior networks for object
detection,''
\newblock {\em Proc. IEEE Conference on Computer Vision and Pattern	Recognition}, 2017.



\bibitem{Krizhevsky_AlexNet_NIPS_2012}
A.~Krizhevsky, I.~Sutskever, and G.~E. Hinton, ``Imagenet classification with deep convolutional neural networks,''
\newblock {\em Proc. Advances in Neural Information Processing Systems}, 2012.


\bibitem{He_SPP_ECCV_2014}
K.~He, X.~Zhang, S.~Ren, and J.~Sun,
\newblock ``Spatial pyramid pooling in deep convolutional networks for visual
recognition,''
\newblock {\em Proc. European Conference on Computer Vision}, 2014.


\bibitem{He_ResNet_CVPR_2016}
K.~He, X.~Zhang, S.~Ren, and J.~Sun, ``Deep residual learning for image recognition,''
\newblock {\em Proc. IEEE Conference on Computer Vision and Pattern Recognition}, 2016.


\bibitem{He_MaskRCNN_ICCV_2017}
K.~He, G.~Gkioxari, P.~Doll{\'a}r, and R.~Girshick,
\newblock ``Mask r-cnn,''
\newblock {\em Proc. IEEE International Conference on Computer Vision}, 2017.

\bibitem{Huang_DenseNet_CVPR_2017}
G.~Huang, Z.~Liu, L.~van~der Maaten, and K.~Q. Weinberger,
\newblock ``Densely connected convolutional networks,''
\newblock {\em Proc. IEEE Conference on Computer Vision and Pattern Recognition}, 2017.

\bibitem{Huang_ApooloScape_TPAMI_2019}
X. Huang, P. Wang, X. Cheng, D. Zhou, Q. Geng, and R. Yang,
``The ApolloScape open dataset for autonomous driving and its application,''
\newblock {\em IEEE Trans. Pattern Analysis and Machine Intelligence}, 2019.


\bibitem{Huang_R2NMS_CVPR_2020}
X. Huang, Z. Ge, Z. Jie, and O. Yoshie,
\newblock ``NMS by representative region: Towards crowded pedestrian detection by proposal pairing,''
\newblock {\em Proc. IEEE Conference on Computer Vision and Pattern Recognition}, 2020.
	

\bibitem{Jain_FDDB_TR_2010}
V. Jain and E. Learned-Miller,
``FDDB: A benchmark for face detection in unconstrained settings,''
\newblock {\em Technical report, University of Massachusetts}, 2010.

\bibitem{Jiang_IoUNet_ECCV_2018}
B.~Jiang, R.~Luo, J.~Mao, T.~Xiao, and Y.~Jiang,
\newblock ``Acquisition of localization confidence for accurate object detection,''
\newblock {\em Proc. European Conference on Computer Vision}, 2018.

\bibitem{Kuznetsova_OpenImage_arXiv_2018}
A. Kuznetsova, H. Rom, N. Alldrin, J. Uijlings, I. Krasin, J. Pont-Tuset, S. Kamali, S. Popov, M. Malloci, T. Duerig, and V. Ferrari,
``The Open Images Dataset V4: Unified image classification, object detection, and visual relationship detection at scale,''
\newblock {\em arXiv:1811.00982}, 2018.


\bibitem{Law_CornerNet_CVPR_2019}
Hei Law and Jia Deng,
``CornerNet: Detecting objects as paired keypoints,''
\newblock {\em Proc. IEEE Conference on Computer Vision and Pattern Recognition}, 2019.

\bibitem{Li_TridentNet_ICCV_2019}
Y. Li, Y. Chen, N. Wang, and Z. Zhang,
\newblock ``Scale-Aware Trident Networks for Object Detection,''
\newblock {\em Proc. IEEE International Conference on Computer Vision}, 2018.


\bibitem{Lin_COCO_ECCV_2014}
T.-Y. Lin, M.~Maire, S.~Belongie, J. Hays, P. Perona, D. Ramanan,  P.~Doll{\'a}r, and L. Zitnick,
\newblock ``Microsoft coco: Common objects in context,''
\newblock {\em Proc. European Conference on Computer Vision}, 2014.

\bibitem{Lin_FPN_CVPR_2017}
T.~Lin, P.~Doll{\'a}r, R.~Girshick, K.~He, B.~Hariharan, and S.~Belongie, ``Feature pyramid networks for object detection,''
\newblock {\em Proc. IEEE Conference on Computer Vision and Pattern Recognition}, 2017.

\bibitem{Lin_Focal_ICCV_2017}
T.~Lin, P.~Goyal, R.~Girshick, K.~He, and P.~Doll{\'a}r, 
``Focal loss for dense object detection,''
\newblock {\em Proc. IEEE International Conference on Computer Vision}, 2017.

\bibitem{Lin_GDFL_ECCV_2018}
C. Lin, J. Lu, G. Wang, and J. Zhou,
\newblock ``Graininess-Aware Deep Feature Learning for Pedestrian Detection,''
\newblock {\em Proc. European Conference on Computer Vision}, 2018.

\bibitem{Liu_SSD_ECCV_2016}
W.~Liu, D.~Anguelov, D.~Erhan, C.~Szegedy, S.~Reed, C.-Y. Fu, and A.~C. Berg,
\newblock ``Ssd: Single shot multibox detector,''
\newblock {\em Proc. European Conference on Computer Vision}, 2016.



\bibitem{Liu_HSFD_CVPR_2019}
W. Liu, S. Liao, W. Ren, W. Hu, and Y. Yu,
``High-level semantic feature detection: A new perspective for pedestrian detection,''
\newblock {\em Proc. IEEE Conference on Computer Vision and Pattern Recognition}, 2019.

\bibitem{Liu_AdaptiveNMS_CVPR_2019}
S. Liu, D. Huang, and Y. Wang,
``Adaptive NMS: Refining pedestrian detection in a crowd,''
\newblock {\em Proc. IEEE Conference on Computer Vision and Pattern Recognition}, 2019.


\bibitem{Mao_WHPD_CVPR_2017}
J. Mao, T. Xiao, Y. Jiang, and Z. Cao,
``What can help pedestrian detection?,''
\newblock {\em Proc. IEEE Conference on Computer Vision and Pattern Recognition}, 2017.

\bibitem{Najibi_AutoFocus_ICCV_2019}
M. Najibi, B. Singh, and L. S. Davis,
``AutoFocus: Efficient multi-scale inference,''
\newblock {\em Proc. IEEE International Conference on Computer Vision}, 2019.

\bibitem{Neuhold_Mapillary_ICCV_2017}
G. Neuhold, T. Ollmann, S. Rota Bulo, and P. Kontschieder,
``The Mapillary vistas dataset for semantic understanding of street scenes,''
\newblock {\em Proc. IEEE International Conference on Computer Vision}, 2017.


\bibitem{Nie_EFGRNet_ICCV_2019}
J. Nie, R. M. Anwer, H.Cholakkal, F. Shahbaz Khan, Y. Pang, and L. Shao, 
``Enriched Feature Guided Refinement Network for Object Detection,''
\newblock {\em Proc. IEEE International Conference on Computer Vision}, 2019.

\bibitem{Ojala_LBP_PAMI_2002}
T. Ojala, M. Pietikainen, and T. Maenpaa,
``Multiresolution gray-scale and rotation invariant texture classification with local binary patterns,''
\newblock {\em IEEE Trans. Pattern Analysis and Machine Intelligence}, vol. 24, no. 7, pp. 971-987, 2002.


\bibitem{Pang_MGAN_ICCV_2019}
Y. Pang, J. Xie, M. H. Khan, R. M. Anwer, F. Shahbaz Khan, and L. Shao, 
``Mask-guided attention network for occluded pedestrian detection,''
\newblock {\em Proc. IEEE International Conference on Computer Vision}, 2019.


\bibitem{Pang_JCS_TIFS_2019}
Y. Pang, J. Cao, J. Wang, and J. Han,
\newblock ``JCS-Net: Joint classification and super-resolution network for small-scale pedestrian detection in surveillance images,''
\newblock {\em IEEE Trans. Information Forensics and Security}, vol. 14, no. 12, pp. 3322-3331, 2019.


\bibitem{Russakovsky_ImageNet_IJCV_2015}
O.~Russakovsky, J.~Deng, H.~Su, J.~Krause, S.~Satheesh, S.~Ma, Z.~Huang, A.~Karpathy, A.~Khosla, M.~Bernstein, A.~C. Berg, and L.~Fei-Fei,
\newblock ``Imagenet large scale visual recognition challenge,''
\newblock {\em International Journal of Computer Vision}, vol. 115, no. 3, pp. 211-252, 2015.



\bibitem{Shao_CrowdHuman_arxiv_2018}
S. Shao, Z. Zhao, B. Li, T. Xiao, G. Yu, X. Zhang, and J. Sun,
``CrowdHuman: A benchmark for detecting human in a crowd,''
\newblock {\em arXiv:1805.00123}, 2018.

\bibitem{Simonyan_VGG_arxiv_2014}
K.~Simonyan and A.~Zisserman, ``Very deep convolutional networks for large-scale image recognition,''
\newblock {\em Proc. International Conference on Learning Representations}, 2015.


\bibitem{Singh_SNIP_CVPR_2018}
B.~Singh and L.~S. Davis,
\newblock ``An analysis of scale invariance in object detection -- snip,''
\newblock {\em Proc. IEEE Conference on Computer Vision and Pattern Recognition}, 2018.

\bibitem{Singh_SNIPER_NIPS_2018}
B.~Singh, M.~Najibi, and L.~S. Davis,
\newblock ``Sniper: Efficient multi-scale training,''
\newblock {\em Proc. Advances in Neural Information Processing Systems}, 2018.



\bibitem{Song_TLL_ECCV_2018}
T. Song, L. Sun, D. Xie, H. Sun, and S. Pu,
\newblock ``Small-scale pedestrian detection based on topological line localization and temporal feature aggregation,''
\newblock {\em Proc. European Conference on Computer Vision}, 2018.

\bibitem{Tian_FCOS_ICCV_2019}
Z. Tian, C. Shen, H. Chen, and T. He,
``FCOS: Fully convolutional one-stage object detection,''
\newblock {\em Proc. IEEE International Conference on Computer Vision}, 2019.

\bibitem{Uijlings_SS_IJCV_2013}
J.~R.~R. Uijlings, K.~E.~A. van~de Sande, T.~Gevers, and A.~W.~M. Smeulders,
\newblock ``Selective search for object recognition,"
\newblock {\em International Journal of Computer Vision}, vol. 104, no. 2, pp. 154-171, 2013.

\bibitem{Viola_RoFace_IJCV_2004}
P.~Viola and M.~Jones,
\newblock ``Robust real-time face detection,"
\newblock {\em International Journal of Computer Vision}, vol. 57, no. 2, pp. 137-154, 2004.

\bibitem{Wang_HOGLBP_ICCV_2009}
X. Wang, T. Han, and S. Yan, 
``An HOG-LBP human detector with partial occlusion handling,''
\newblock {\em Proc. IEEE International Conference on Computer Vision}, 2009.

\bibitem{Wang_Repulsion_CVPR_2018}
X. Wang, T. Xiao, Y. Jiang, S. Shao, J. Sun, and C. Shen, 
``Repulsion loss: Detecting pedestrians in a crowd,''
\newblock {\em Proc. IEEE Conference on Computer Vision and Pattern Recognition}, 2018.


\bibitem{Wang_LRF_ICCV_2019}
T. Wang, R. M. Anwer, H. Cholakkal, F. S. Khan, Y. Pang, and L. Shao, 
``Learning rich features at high-speed for single-shot object detection,''
\newblock {\em Proc. IEEE International Conference on Computer Vision}, 2019.

\bibitem{Wu_TCED_CVPR_2020}
J. Wu, C. Zhou, M. Yang, Q. Zhang, Y.Li, and J. Yuan, 
``Temporal-context enhanced detection of heavily occluded pedestrians,''
\newblock {\em Proc. IEEE Conference on Computer Vision and Pattern Recognition}, 2020.

\bibitem{Wu_SelfMimic_ACM_2020}
J. Wu, C. Zhou, Q. Zhang, M. Yang, and J. Yuan, 
``Self-Mimic Learning for Small-scale Pedestrian Detection,''
\newblock {\em Proc. ACM International Conference on Multimedia}, 2020.

\bibitem{Xiao_JDIFL_CVPR_2017}
T. Xiao, S. Li, B. Wang, L. Lin, and X. Wang,
Joint Detection and Identification Feature Learning for Person Search
\newblock {\em Proc. IEEE Conference on Computer Vision and Pattern Recognition}, 2017.

\bibitem{Ye_DGC_TIP_2019}
M. Ye, J. Li, A. J Ma, L. Zheng, and P. Yuen,
``Dynamic graph co-matching for unsupervised video-based person re-identification,''
\newblock {\em IEEE Trans. Image Processing}, vol. 28, no. 6, pp. 2976-2990, 2019.



\bibitem{Ye_BDCC_TIFS_2019}
M. Ye, X. Lan, Z. Wang, and P. Yuen,
``Bi-directional center-constrained top-ranking for visible thermal person re-identification,''
\newblock {\em IEEE Trans. Information Forensics and Security}, 10.1109/TIFS.2019.2921454, 2019.



\bibitem{Yang_WiderFace_CVPR_2016}
S. Yang, P. Luo, C. Change Loy, and X. Tang,
``WIDER FACE: A face detection benchmark,''
\newblock {\em Proc. IEEE Conference on Computer Vision and Pattern Recognition}, 2016.

\bibitem{Yang_PascalFace_IVC_2014}
J. Yan, X. Zhang, Z. Lei, and S. Z. Li,
``Face detection by structural models,'' 
\newblock {\em Image and Vision Computing}, 2014


\bibitem{Yang_RepPoint_ICCV_2019}
Z. Yang, S. Liu, H. Hu, L. Wang, and S. Lin, 
``RepPoints: Point set representation for object detection,''
\newblock {\em Proc. IEEE International Conference on Computer Vision}, 2019.


\bibitem{Yu_BDD100K_arXiv_2019}
F. Yu, W. Xian, Y. Chen, F. Liu, M. Liao, V. Madhavan, and T. Darrell,
``BDD100K: A diverse driving video database with scalable annotation tooling,''
\newblock {\em arXiv:1805.04687}, 2019.


\bibitem{Zhang_FCF_CVPR_2015}
S. Zhang, R. Benenson, and B. Schiele,
``Filtered Channel Features for Pedestrian Detection,''
\newblock {\em Proc. IEEE Conference on Computer Vision and Pattern Recognition}, 2015.


\bibitem{Zhang_CityPersions_CVPR_2017}
S.~Zhang, R.~Benenson, and B.~Schiele, 
``Citypersons: A diverse dataset for pedestrian detection,''
\newblock {\em Proc. IEEE Conference on Computer Vision and Pattern Recognition}, 2017.

\bibitem{Zhang_FasterATT_CVPR_2018}
S.~Zhang, J. Yang, and B.~Schiele, 
``Occluded Pedestrian Detection Through Guided Attention in CNNs,''
\newblock {\em Proc. IEEE Conference on Computer Vision and Pattern Recognition}, 2017.



\bibitem{Zhang_RefineDet_CVPR_2018}
S.~Zhang, L.~Wen, X.~Bian, Z.~Lei, and S.~Z. Li,
\newblock ``Single-shot refinement neural network for object detection,''
\newblock {\em Proc. IEEE Conference on Computer Vision and Pattern Recognition}, 2018.


\bibitem{Zhang_ORCNN_ECCV_2018}
S.~Zhang, L.~Wen, X.~Bian, Z.~Lei, and S.~Z. Li,
\newblock ``Occlusion-aware R-CNN: Detecting pedestrians in a crowd,''
\newblock {\em Proc. European Conference on Computer Vision}, 2018.


\bibitem{Zhang_DES_CVPR_2018}
Z.~Zhang, S.~Qiao, C.~Xie, W.~Shen, B.~Wang, and A.~L. Yuille,
\newblock ``Single-shot object detection with enriched semantics,''
\newblock {\em Proc. IEEE Conference on Computer Vision and Pattern Recognition}, 2018.

\bibitem{Zhao_M2Det_AAAI_2018}
Q.~Zhao, T.~Sheng, Y.~Wang, Z.~Tang, Y.~Chen, L.~Cai, and H.~Ling,
\newblock ``M2det: A single-shot object detector based on multi-level feature
pyramid network,''
\newblock {\em Proc. AAAI Conference on Artificial Intelligence}, 2018.

\bibitem{Zhou_CenterNet_arXiv_2019}
X. Zhou, D. Wang, and P. Krähenbühl,
``Objects as points,''
\newblock {\em arXiv:1904.07850}, 2019.

\bibitem{Zhou_DFT_ICCV_2019}
C. Zhou, M. Yang, and J. Yuan, 
``Discriminative feature transformation for occluded pedestrian detection,''
\newblock {\em Proc. IEEE International Conference on Computer Vision}, 2019.

\bibitem{Zhou_Bibox_ECCV_2018}
C. Zhou and J. Yuan, 
``Bi-box regression for pedestrian detection and occlusion estimation,''
\newblock {\em Proc. European Conf. on Computer Vision}, 2018.

\bibitem{Zhou_MLL_ICCV_2017}
C. Zhou and J. Yuan, 
``Multi-label learning of part detectors for heavily occluded pedestrian detection,''
\newblock {\em Proc. IEEE International Conference on Computer Vision}, 2017.


\bibitem{Zhu_AFW_CVPR_2012}
X. Zhu and D. Ramanan,
``Face detection, pose estimation, and landmark localization in the wild,''
\newblock {\em Proc. IEEE Conference on Computer Vision and Pattern Recognition}, 2012.


\end{thebibliography}
\end{document}